\documentclass{article}

\usepackage[preprint]{corl_2022} % Use this for the initial submission.

\usepackage{caption}
\usepackage{subfig}
\usepackage{wrapfig}

\usepackage{algorithmicx}
\usepackage{algorithm}
\usepackage{algpseudocode}

\usepackage{amsmath}
\usepackage{amssymb}
\usepackage{amsfonts}
\usepackage{amsthm}
\usepackage{amsbsy}
\usepackage{mathrsfs}
\usepackage{mathtools}

\usepackage{booktabs}
\usepackage{array}
\usepackage{makecell}

\usepackage{multirow}

\newcommand*\diff{\mathop{}\!\mathrm{d}}

\newcommand{\sampler}{{\text{sampler}}}

\usepackage{xcolor}
\definecolor{darkblue}{rgb}{0.0,0.0,0.3}
\definecolor{commentBlue}{rgb}{0.0,0.0,0.8}

\definecolor{blau}{RGB}{0, 69, 134}
\definecolor{orangeR}{RGB}{255,140,0}
\definecolor{gruen}{RGB}{0,150,0}
\definecolor{lila}{rgb}{0.49400,0.18400,0.55600}
\definecolor{rot}{rgb}{1,0.12500,0.009800}
\definecolor{hellblau}{RGB}{23, 190, 207}
\definecolor{rosa}{RGB}{207, 23, 190}

\def\generatePlots{0}

\usepackage{tikz}
\usepackage{pgfplots}
\usetikzlibrary{external}
\if\generatePlots1
\fi

\newcounter{tikzCounter}
\newcommand{\includetikz}[1]{%
	\if\generatePlots1
		\tikzsetnextfilename{tikz/tikz-\thetikzCounter}%
		\input{#1}%
	\else
		\includegraphics{tikz/tikz-\thetikzCounter}
	\fi
	\addtocounter{tikzCounter}{1}
}\input

\pgfplotsset{compat=newest}
\usetikzlibrary{plotmarks}
\usepackage{grffile}
\usepgfplotslibrary{statistics}

\pgfplotsset{
	legend image code/.code={
		\draw[mark repeat=2,mark phase=2]
		plot coordinates {
			(0cm,0cm)
			(0.15cm,0cm)        %% default is (0.3cm,0cm)
			(0.3cm,0cm)         %% default is (0.6cm,0cm)
		};%
	}
}
\pgfplotsset{label style={font=\scriptsize},
	tick label style={font=\scriptsize} }

\usetikzlibrary{3d}
\usetikzlibrary{shapes, arrows, fit, calc, positioning, matrix}
\usetikzlibrary{decorations.markings}

\tikzset{font=\footnotesize}

\title{Learning Multi-Object Dynamics with Compositional Neural Radiance Fields}

\author{
	Danny Driess\\
	TU Berlin
	\And
	Zhiao Huang\\
	UC San Diego
	\And 
	Yunzhu Li\\
	MIT
	\And 
	Russ Tedrake\\
	MIT
	\And
	Marc Toussaint\\
	TU Berlin
}

\begin{document}
\maketitle
	
\vspace{-0.2cm}
\begin{abstract}
We present a method to learn compositional multi-object dynamics models from image observations based on implicit object encoders, Neural Radiance Fields (NeRFs), and graph neural networks.
NeRFs have become a popular choice for representing scenes due to their strong 3D prior.
However, most NeRF approaches are trained on a single scene, representing the whole scene with a global model, making generalization to novel scenes, containing different numbers of objects, challenging.
Instead, we present a compositional, object-centric auto-encoder framework that maps multiple views of the scene to a \emph{set} of latent vectors representing each object separately.
The latent vectors parameterize individual NeRFs from which the scene can be reconstructed.
Based on those latent vectors, we train a graph neural network dynamics model in the latent space to achieve compositionality for dynamics prediction.
A key feature of our approach is that the latent vectors are forced to encode 3D information through the NeRF decoder, which enables us to incorporate structural priors in learning the dynamics models, making long-term predictions more stable compared to several baselines.
Simulated and real world experiments show that our method can model and learn the dynamics of compositional scenes including rigid and deformable objects.\\
Video: \url{https://dannydriess.github.io/compnerfdyn/}
\end{abstract}


\section{Introduction}\label{sec:introduction}
\vspace{-0.3cm}
Learning models from observations that predict the future state of a scene is a fundamental concept for enabling an agent to reason about actions to achieve a desired goal.
A major challenge in learning predictive models is that raw observations such as images are usually high-dimensional.
Therefore, a common approach is to map the observation space into a lower-dimensional latent representation of the scene via an auto-encoder structure.
Based on those latent vectors, a dynamics model can be learned that predicts the next latent state, conditioned on actions an agent takes.
An intuition for this is that if a latent vector is sufficient to reconstruct the observations, then it contains enough information about the scene to learn a dynamics model on top of it.
While an auto-encoder structure combined with a latent dynamics model is a general approach that is applicable for a large variety of tasks, it raises multiple challenges.
First, scenes in our world are \emph{composed} of multiple objects.
Therefore, a fixed-size latent vector has difficulties in generalizing over different and changing numbers of objects in the scene than during training, both due to the limited capacity of fixed-size vectors and lack of diversity in the training distribution.
Second, image observations are 2D, but the 3D structure of our world is essential for many tasks to reason about the underlying physical processes governing the dynamics the model should predict.
Dealing with occlusions, object permanence, and ambiguities in 2D views is challenging for 2D image representations.
Importantly, many forward predictive models in visual observation spaces suffer from instabilities in making long-term predictions, often manifested in blurry image predictions \cite{ebert2018visual}.
\begin{figure}
    \captionsetup[subfloat]{captionskip=2pt}
    \tikzset{external/export next=false}
    \begin{tikzpicture}[font=\scriptsize]
    \node[] at (0,0) (a) {$t=0$};
    \node[] at (9,0) (b) {$t=15$};
    \draw[->] (a) -- (b);
    \end{tikzpicture}\\[-0.6cm]
    \centering
    \subfloat[Bottom row renderings of forward predictions with dynamic model, top row ground truth]{
    	\shortstack{
	    	\includegraphics[trim={1cm 1.1cm 1cm 0}, clip, height=1.35cm]{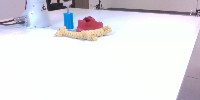}
	    	\includegraphics[trim={1cm 1.1cm 1cm 0}, clip, height=1.35cm]{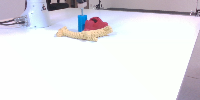}
	    	\includegraphics[trim={1cm 1.1cm 1cm 0}, clip, height=1.35cm]{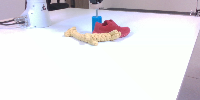}
	    	\includegraphics[trim={1cm 1.1cm 1cm 0}, clip, height=1.35cm]{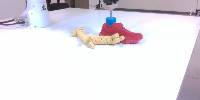}\\
	    	\includegraphics[trim={1cm 1.1cm 1cm 0}, clip, height=1.35cm]{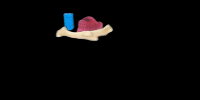}
	    	\includegraphics[trim={1cm 1.1cm 1cm 0}, clip, height=1.35cm]{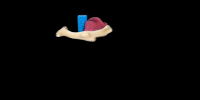}
	    	\includegraphics[trim={1cm 1.1cm 1cm 0}, clip, height=1.35cm]{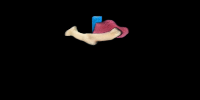}
	    	\includegraphics[trim={1cm 1.1cm 1cm 0}, clip, height=1.35cm]{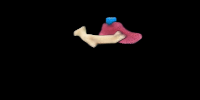}
	    }
    	\label{fig:firstPage:pred}
	}
	\subfloat[novel view]{
		\shortstack{
			\includegraphics[trim={4cm 1cm 5cm 1cm}, clip,height=1.5cm]{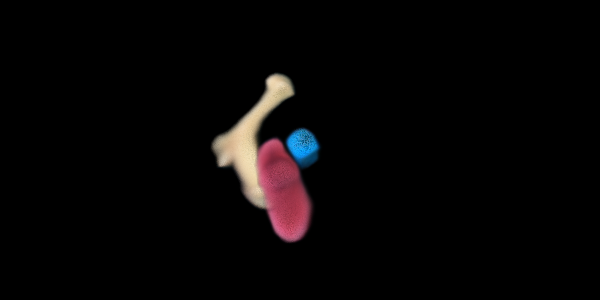}\\
			\includegraphics[trim={5cm 1cm 4cm 1cm}, clip,height=1.5cm]{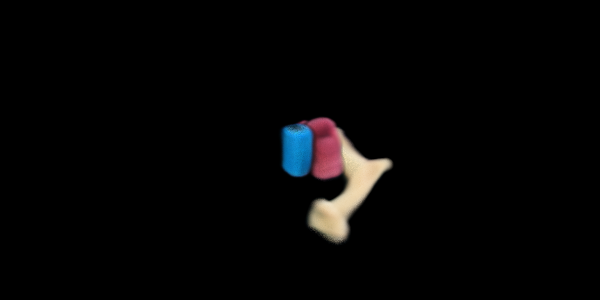}
		}
		\label{fig:firstPage:novel}
	}
    \vspace{-0.2cm}
    \caption{\small Visual forward predictions with our model. (a) left: initial scene, (a) right: after 15 prediction steps into the future. (b) renderings of the model after 15 prediction steps from novel views. Future predictions are based on the initial observation of the scene at $t=0$ and then rendered from the prediction of the latent vectors. Blue pusher is articulated by the robot. Despite multiple objects interacting, predictions are sharp and accurate.\vspace{-0.6cm}}
    \label{fig:firstPage}
\end{figure}

One way to address these issues is to incorporate inductive biases and structural priors in the model architectures.
Li~et~al.~\cite{li20223d} proposed to use Neural Radiance Fields (NeRFs) \cite{mildenhall2020nerf} as a decoder within an auto-encoder to learn dynamics models in latent spaces.
NeRFs exhibit strong structural priors about the 3D world, leading to increased performance over 2D baselines.
However, the approach of \cite{li20223d} represents the whole scene as a single latent vector, which we found insufficient for scenes that are composed of multiple, different numbers of objects, both in terms of representation and dynamics prediction.

In the present work, we aim to overcome these challenges by incorporating inductive biases on the compositional nature and underlying 3D structure of our world both in learning the latent representations themselves and the dynamics model.
We propose a compositional, object-centric auto-encoder framework whose latent vectors are used to learn a compositional forward dynamics model in that learned latent space based on graph neural networks (GNN).
More specifically, we learn an implicit object encoder that maps image observations of the scene from multiple views to a set of latent vectors that each represent an object in the scene separately.
These latent object encodings then parameterize individual NeRFs for each object.
We apply compositional rendering techniques to synthesize images from multiple viewpoints, which forces the object-centric NeRF functions and the corresponding latent vectors to learn precise 3D configurations of the constituting objects.
This 3D inductive bias both in the encoder and the compositional NeRF decoder enables us to incorporate priors from the models' own predictions about objects interactions via an estimated adjacency matrix into learning the GNN dynamics model, making long-term dynamics predictions more stable.
This long term-stability allows us utilize a planning method based on RRTs in the latent space.

In our evaluations, we show through comparisons that non-compositional auto-encoder frameworks and non-compositional dynamics models struggle with tasks containing multiple objects, while our framework generalizes well over different numbers of objects than during training and is capable of generating sharp and stable long-term predictions. 
Relative to more traditional multibody system identification \cite{tedrake2022}, these models learn the geometry of unknown objects in addition to (implicitly) learning the inertial and contact parameters.
We demonstrate the performance of the approach in terms of image reconstruction error, dynamics prediction error, and planning, generalizing over different numbers of objects than during training.
Our experiments include rigid and deformable objects in simulation and with a real robot.
To summarize, our main contributions are
\vspace{-0.2cm}
\begin{itemize}
    \item A compositional scene encoding framework that uses implicit object encoders and NeRF decoders for each object, forcing the view-invariant latent representation to learn about the 3D structure of the problem in a composable way.
    \vspace{-0.1cm}
    \item A factored dynamics model in the latent space as a graph neural network (GNN), exploiting the compositional nature of the scene representation and an adaptive adjacency matrix estimated from the model itself to yield stable long-term predictions.
\end{itemize}

\vspace{-0.3cm}
\section{Related Work}\label{sec:relatedWork}
\vspace{-0.3cm}

\textbf{Learning Dynamics Models for Compositional Systems.}
Graph neural networks (GNNs) have shown great promise in introducing relational inductive biases~\cite{battaglia2018relational}, enabling them to model the dynamics of compositional systems consisting of interactions between multiple objects~\cite{battaglia2016interaction,chang2016compositional,li2019propagation,sanchez2018graph,funk2022learn2assemble,silver2020planning}, large-scale dynamical systems represented using particles and meshes~\cite{mrowca2018flexible,li2018learning,li2020visual,ummenhofer2019lagrangian,sanchez2020learning,pfaff2020learning}, or from visual observations \cite{ye2019compositional,hsieh2018learning,watters2017visual,yi2019clevrer,qi2020learning,tung20203d,NEURIPS2018_713fd63d}.
Our method differs from prior work by learning compositional scene representations grounded in 3D space directly from visual observations. Our novel combination of implicit object encoders and graph-based neural dynamics models reflects the structure of the underlying scene, which endows our agent with better generalization ability in handling complicated compositional dynamic environments.

\textbf{NeRF for Compositional and Dynamic Scenes.}
Recent advances on neural implicit representations \cite{xie2021neural} have demonstrated widespread success in image synthesis or 3D reconstruction~\cite{mescheder2019occupancy,peng2020convolutional,park2019deepsdf,sitzmann2019scene,saito2019pifu}. Notably, Neural Radiance Fields (NeRF) show impressive results on novel-view synthesis~\cite{mildenhall2020nerf}.
Initial NeRF approaches were trained on a single scene without generalization.
Prior work \cite{guo2020object,stelzner2021decomposing,ost2021neural,yu2021unsupervised,elich2020weakly} have since proposed to modify neural scene representations to make them compositional for static scenes without considering dynamics of object interactions.
People have also extended NeRF to enable view synthesis from a sparse set of views~\cite{yu2021pixelnerf}, as well as modeling dynamic scenes by learning implicitly represented flow fields or time-variant latent codes~\cite{pumarola2021d,park2021nerfies,du2021neural,xu2021h,park2021hypernerf,niemeyer2019occupancy,tretschk2021non,li2021neural,xian2021space,ost2021neural,li2021neurala}.
However, these approaches for dynamic environments typically interpolate over a single time sequence and are not able to handle scenes of different initial configurations or different action sequences, limiting their use in downstream planning and control tasks.
Li~et~al.~\cite{li20223d} addressed this issue by combining an NeRF auto-encoding framework with modeling the dynamics in a latent space.
Yet, they employed a single latent vector as the whole scene representation, which we will show is insufficient at modeling compositional systems. 
In contrast, our method considers a graph-based scene representation to capture the structure of the underlying scene and achieves significantly better generalization performance than~\cite{li20223d}.

\textbf{Implicit Models in Robotics.}
Implicit models in robotics have been explored, e.g., for grasping \cite{breyer2020volumetric,jiang2021synergies,van2020learning,ichnowski2021dex} or more general manipulation planning constraints \cite{driess2022CoRL, ha2021learning}.
Analytic signed distance functions \cite{hauser1,pfrommer2020contactnets,driess2022CoRL} or learned NeRFs \cite{robotNavigtationInNeRF2022} are used for trajectory planning.
One assumption in \cite{driess2022CoRL} and \cite{ha2021learning} is that signed-distance values are available during training.
Our work, in contrast, directly operates on RGB images without requiring explicit 3D shape supervision.

\vspace{-0.1cm}
\section{Overview -- Compositional Visual Dynamics Learning}
\vspace{-0.3cm}

\begin{figure}
	\vspace{-0.3cm}
	\centering
	\scalebox{0.75}{\input{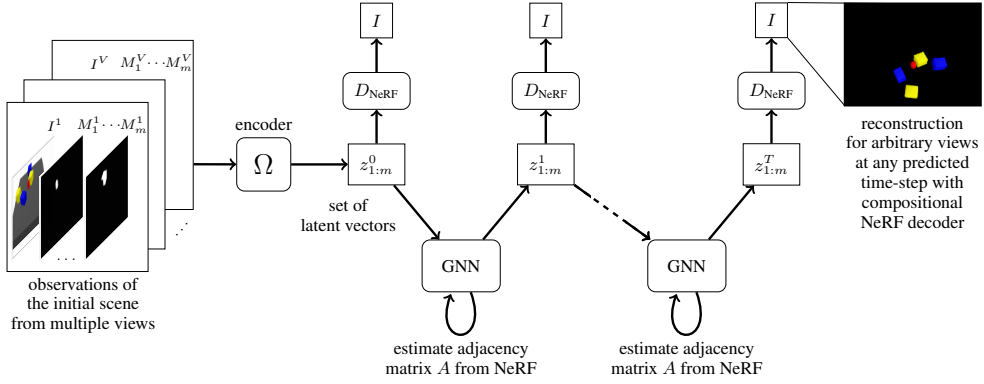}}
	\caption{\small Overview of the dynamics prediction framework. The initial scene observations are encoded with $\Omega$ into a set of latent vectors $z_{1:m}$, each representing the objects individually. The GNN dynamics model predicts the evolution of the latent vectors. At each step, the predicted latent vectors can be rendered into an arbitrary view with the compositional NeRF decoder.
	Refer to the appendix for visualizations of $\Omega$ and the GNN.}
	\vspace{-0.3cm}
	\label{fig:overview}
\end{figure}

Our dynamics learning framework (Fig.~\ref{fig:firstPage}) consists of three parts, an object encoder $\Omega$ turning observations into a set of latent vectors $z_{1:m}$, a compositional NeRF-based decoder $D_\text{NeRF}$ that renders the latent vectors back into images of the scene to train the encoder, and a graph neural network dynamics model $F_\text{GNN}$ predicting the evolution of the scene in the latent space.
This section gives a high-level overview, while Sec.~\ref{sec:CICNerf}, Sec.~\ref{sec:GNN} as well as the appendix Sec.~\ref{app:autoencoder}, Sec.~\ref{app:gnn} provide details.

Assume that a scene is observed by RGB images $I^i\in\mathbb{R}^{3\times h_I\times w_I}$, $i=1,\ldots, V$ from $V$ many camera views and that the scene contains $m$ objects $j=1,\ldots,m$.
We further assume to have access to the camera projection matrices $K^i\in\mathbb{R}^{3\times 4}$ for each view and binary masks $M_j^i\in\left\{0,1\right\}^{h_I\times w_I}$ of each object $j$ in view $i$.
Given those posed images and masks, the goal is to learn an encoder $\Omega$ that fuses the information of the objects observed from the multiple views into a set of latent vectors $z_{1:m}$ by querying $\Omega$ on the individual masks $M_j^{1:V}$ such that
\begin{align}
z_{j} = \Omega\left(I^{1:V}, K^{1:V}, M^{1:V}_{j}\right) \in\mathbb{R}^k\label{eq:omega}
\end{align}
represents the object $j$ separately. $\Omega$ is trained end-to-end with a NeRF decoder $D_\text{NeRF}$ reconstructing
\begin{align}
I = D_\text{NeRF}(z_{1:m}, K)
\end{align}
for arbitrary views specified by the camera matrix $K$ from the set of latent object representations $z_{1:m}$.
The initial observation of the scene is encoded with $\Omega$ into the initial latent vectors $z_{1:m}^0$.
The GNN dynamics model $z_{1:m}^{t+1} = F_\text{GNN}\left(z_{1:m}^t\right)$ then generates long-term predictions of future latent states $z^t_{1:m}$ that can also be decoded with $D_\text{NeRF}$ to yield visual predictions from arbitrary views.

\section{Encoding Scenes with Compositional Image-Conditioned NeRFs}\label{sec:CICNerf}
\vspace{-0.1cm} 

\subsection{Implicit Object Encoder}\label{sec:implicitObjectEncoder}
\vspace{-0.2cm}
Instead of learning $\Omega$ defined in \eqref{eq:omega} as a direct mapping from images, camera matrices and masks to the latent vectors, we first encode each object in the scene as a feature-valued \emph{function} over 3D space, conditioned on the image observations.
This allows us to incorporate multiple views of the objects in a geometrically consistent way, as well as to apply 3D affine transformations to the objects, which will be important for the dynamics model (Sec.~\ref{sec:GNN}).
This function is then turned into a latent vector by evaluating it on a workspace set followed by a 3D convolutional network.

All object feature functions are based on the \emph{same} feature encoder
$
    E(I^i, K^i(x)) \in\mathbb{R}^{n_o}
$
that outputs an $n_o$-dimensional feature vector from the image $I^i$ of view $i$ at any 3D world coordinate $x\in\mathbb{R}^3$.
This is realized by first projecting $x$ into camera coordinates $K^i(x) = \left(u^i(x), v^i(x), d^i(x)\right)^T \in\mathbb{R}^3$ where $u^i(x), v^i(x)$ are pixel coordinates in the image plane and $d^i(x)\in\mathbb{R}$ is the depth of $x$ from the camera origin.
Hence, $E$ is a function of the camera coordinates only and not of absolute world coordinates.
Using bilinear interpolation, the encoder $E(I^i, K^i(x))$ determines the RGB values of $I^i$ at $(u^i(x),v^i(x))$ which are passed through a dense neural network (MLP).
Parallel to this, a dense MLP encoding of $K^i(x)$ is computed.
The concatenated outputs of both MLPs define the encoding feature vector $E(I^i, K^i(x))$. 
Intuitively, $E(I^i, K^i(x))$ is a feature vector computed from what can be seen of the world at $x$ in the image $I^i$ from viewpoint $i$, taking into account its location relative to the camera origin of the view $i$, which is important not only to enable the model to reason about the 3D geometry, but also to enable us to obtain a functional representation of a specific object $j$. Namely, we define the feature function for object $j$ by summing over the individual views $i$
\begin{align}
    y_j(x) = \frac{1}{p(x)}\sum_{i:~ K^i(x)\in M_j^i} E(I^i, K^i(x)) \in \mathbb{R}^{n_o}~~~~\text{with}~~~~p(x) = \sum_{i:~ K^i(x)\in M_j^i} 1.\label{eq:yObjectFunction}
\end{align}
Importantly, for a specific $x$, this sum only takes those views $i$ into account where the object $j$ can be seen, i.e., where the camera coordinates $K^i(x)$ of $x$ are within the object's mask $M_j^i$.
We define $y_j(x) = 0\in\mathbb{R}^{n_o}$ if $p(x) = 0$, meaning if an object is not observed from any view at $x$, the corresponding feature vector is zero.
An advantage of this formulation is that it naturally handles occlusions in different views and fuses the observations from different views consistently.

Given the implicit object descriptor function $y_j(\cdot)$ of object~$j$, we turn it into a latent vector $z_j\in\mathbb{R}^k$ representing object $j$ with a 3D convolutional network $\Phi$ as follows.
Formally, $z_j = \Phi(y_j)$ is a function of the object \emph{function}.
As discussed in \cite{driess2022CoRL}, learning a function of a function can be realized with neural networks by evaluating $y_j$ on a workspace set.
We assume that the interactions in the scene happen within a workspace set $\mathcal{X}\subset\mathbb{R}^3$ that is large enough to contain all objects.
This workspace set is discretized as the voxel grid $\mathcal{X}_h\in\mathbb{R}^{d\times h\times w}$.
The object descriptor functions are then evaluated on $\mathcal{X}_h$ which produces an object feature voxel grid that is processed with a 3D convolutional neural network leading to the latent vector $z_j\in\mathbb{R}^k$, i.e.\
\begin{align}
    z_j = \Phi(y_j) = \text{CNN}(y_j(\mathcal{X}_h)). \label{eq:Phi}
\end{align}
Note that the same workspace set $\mathcal{X}_h$ is used for all objects. The appendix contains visualizations of the architectures of $E$, $y$ and $\Phi$.

In summary, the object encoder
$
	z_{1:m} = \Omega\left(I^{1:V}, K^{1:V}, M^{1:V}_{1:m}, \mathcal{X}_h\right)
$
maps images from multiple views, object masks and the set $\mathcal{X}_h$ to latent vectors.
The resulting $z_j$'s contain not only the appearance of the objects, but also their spatial configurations in the scene relative to other objects.

\vspace{-0.1cm}
\subsection{Decoder as Compositional, Conditional NeRF Model}\label{sec:decoder}
\vspace{-0.2cm}
The general idea of NeRF \cite{mildenhall2020nerf} is to learn a function $f$ that predicts at a 3D world coordinate $x\in\mathbb{R}^3$ the RGB color value $c(x)\in\mathbb{R}^{3}$ and volume density $\sigma(x)\in\mathbb{R}_{\ge0}$.
Based on $(\sigma(\cdot), c(\cdot)) = f(\cdot)$, images from arbitrary views and camera configurations can be rendered by determining the color of the pixels along corresponding camera rays through volumetric rendering.
For details, see Sec.~\ref{app:autoencoder}.

Compared to this standard NeRF formulation where one single model is used to represent the whole scene, we associate separate NeRFs with each object, meaning that the NeRF for object $j$
\begin{align}
    (\sigma_j(x), c_j(x)) = f_j(x) = f(x, z_j)\label{eq:compNeRFs}
\end{align}
is conditioned on $z_j$ for $j=1,\ldots,m$.
$\sigma_j$ is the density and $c_j$ the color prediction for object $j$, respectively.
To turn those $f_{1:m}$ back into a global NeRF model that can be rendered to an image, we sum the individual predicted object densities $\sigma(x) = \sum_{j=1}^m \sigma_j(x)$ 
and obtain the colors as their density weighted combination $c(x) = \frac{1}{\sigma(x)}\sum_{j=1}^m \sigma_j(x)c_j(x)$.
These composition formulas have been proposed multiple times in the literature, e.g.\ \cite{Niemeyer2020GIRAFFE, stelzner2021decomposing}.
This composition forces the individual NeRFs to learn the 3D configuration of each object individually and therefore ensures that each $f_j$ only predicts the object where it is located in the 3D space.

To summarize, the compositional NeRF-decoder $D_\text{NeRF}$ takes the set of latent vectors $z_{1:m}$ for objects $j=1,\ldots,m$ and the camera matrix $K$ for a desired view as input to render
$
    I = D_\text{NeRF}(z_{1:m}, K).
$
Since we only represent the objects and not the background as NeRFs, rendering the composed NeRF will yield an image with the background subtracted.
In the experiments, we investigate the importance of the decoder being both compositional and a NeRF.

\vspace{-0.1cm}
\subsection{Training}\label{sec:encoderDecoderTraining}
\vspace{-0.2cm}
The auto-encoder framework is trained end-to-end on an $\text{L}_2$ image reconstruction loss for view $i$
\begin{align}
\mathcal{L}^i = \sum_{(u,v)\in\hat{M}_\text{tot}^i}\left\|\left(I^i\circ M_\text{tot}^i\right)_{uv} \!\!\!-\! D_\text{NeRF}\!\left(\Omega\left(I^{1:V}, K^{1:V}, M^{1:V}_{1:m}, \mathcal{X}_h\right), K^i\right)_{uv}\right\|_2^2.
\end{align}
Since solely the objects are represented as NeRFs and not the background, we compute the union of the masks of the individual objects
$
    M_\text{tot}^i = \bigvee_{j=1}^m M_j^i
$
and define the target image as $I^i\circ \hat{M}_\text{tot}^i$ with $\hat{M}_\text{tot}^i$ being a slightly enlarged union mask.
Please refer to the appendix Sec.~\ref{app:autoencoder} for more details.

\vspace{-0.1cm}
\section{Latent Dynamics Model with Graph Neural Networks}\label{sec:GNN}
\vspace{-0.3cm}
Having trained the auto-encoder framework, we learn a graph neural network dynamics model
\begin{align}
z_{1:m}^{t+1} = F_\text{GNN}\left(z_{1:m}^t, A^t\right)\label{eq:gnn}
\end{align}
in the latent space, where $A^t\in\left\{0,1\right\}^{m\times m}$ is the adjacency matrix at time $t$.
Following \cite{li2019propagation}, we use multi-step message passing to deal with cases where multiple objects interact within one prediction step.
Refer to the appendix Sec.~\ref{app:gnn} and Algo.~\ref{algo:forwardPred} for more details about our GNN dynamics model.

\textbf{Adjacency Matrix from Learned Model.}
The adjacency matrix $A$ in the GNN dynamics model \eqref{eq:gnn} plays an important role in indicating which objects interact.
While a dense adjacency matrix, i.e.\ a graph
where each object interacts with all other objects, would in principle work as the GNN could figure out from the latent vectors which objects interact, we found that the long-horizon prediction performance is greatly increased if $A$ is more selective in reflecting which objects actually interact.

We propose to utilize the NeRF decoder density prediction $\sigma_j$ for each object to determine the adjacency matrix from the models' own predictions during training and planning. 
In order to do so, we define the entries of the adjacency matrix between objects $i$ and $j$ based on the collision integral
\begin{align}
	A_{ij} = \begin{cases}
	1 & \int_{\mathcal{X}}[\sigma(x, z_i)>\kappa][\sigma(x, z_j)>\kappa]\diff x > 0\\
	0 & \text{else}
	\end{cases}
\end{align}
over the density predictions of the learned NeRF model for a threshold $\kappa \ge 0$.
Estimating $A$ this way takes the actual 3D geometry of the objects in the scene into account and thereby informs the GNN dynamics model, leading to more stable predictions.
Please refer to the appendix Sec.~\ref{app:gnn} for more details about $A$ and how it is used in the forward prediction Algo.~\ref{algo:forwardPred}.

\textbf{Actions.}
So far, we have formulated the GNN dynamics model without a notion of actions.
We interpret an action as an intervention to a latent vector and train the GNN to predict the latent vectors at the next time step as a result to this modification.
This allows us to not explicitly distinguish between controlled and uncontrolled/passive objects.
In order to realize these interventions and hence to incorporate actions in the first place, we utilize the fact that our object encoder is built from an implicit representation.
Assume that an action is a rigid transformation $q\in\mathbb{R}^7$ applied on object $j$.
As described in Sec.~\ref{app:autoencoder} we can modify the object's latent vector $z_j^t$ into the transformed $\bar{z}^t_j =  z_j^{t+1} = \mathcal{T}(q)[z_j]$ representing the rigidly transformed object $j$.
The model $F_\text{GNN}$ then predicts how the other objects in the scene react to this rigid transformation of the articulated object.

\section{Experiments}
\vspace{-0.3cm}
We demonstrate our framework on pushing tasks in scenes containing multiple objects both in simulation and in the real world.
For a quantitative analysis and comparison to multiple baselines, we investigate here the forward prediction error of the model in the image space rendered from the learned model over long-horizons.
Please refer to the supplementary video showing the reconstructions of the model, novel scene generation, forward predictions and planning/execution results as well as the appendix for more details and further experiments.
Our scenarios are challenging, as they are composed of multiple, interacting objects, sparse rewards, and complex dynamics \cite{hogan2016feedback, zhou2019pushing, 21-driess-IJRR, schubert2021learning}.

\vspace{-0.1cm}
\subsection{Visual Reconstruction and Prediction Performance -- Comparison to Baselines}
\vspace{-0.2cm}
We compare our framework to non-compositional scene representations, non-compositional dynamics models, 2D CNN baselines (visual foresight) without NeRF as decoder, and the importance of estimating the adjacency matrix from the model itself.

\begin{figure}
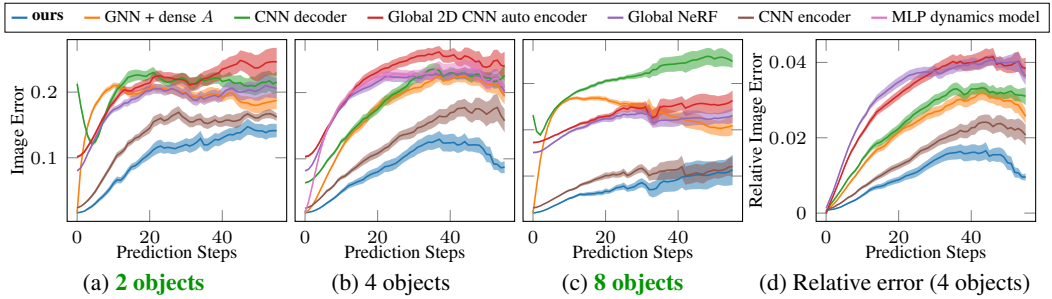

	\centering
	\vspace{-0.2cm}
	\scalebox{0.7}{\input{plots/baselinesLegend}}\\[-0.3cm]
	\captionsetup[subfloat]{captionskip=0pt}
	\subfloat[\color{gruen}\textbf{2 objects}]{%
		\hspace{-0.2cm}
		\input{plots/baselinesErrorPredTrueEMask_nM_2}%
		\label{fig:baselinesImage:2}
		\hspace{-0.4cm}
	}%
	\subfloat[4 objects]{%
		\input{plots/baselinesErrorPredTrueEMask_nM_4}%
		\label{fig:baselinesImage:4}
		\hspace{-0.4cm}
	}%
	\subfloat[\color{gruen}\textbf{8 objects}]{%
		\input{plots/baselinesErrorPredTrueEMask_nM_8}%
		\label{fig:baselinesImage:8}
		\hspace{-0.3cm}
	}%
	\subfloat[Relative error (4 objects)]{%
		\input{plots/baselinesErrorPredRecon_nM_4}%
		\label{fig:baselinesPredRecon}
	}%
	\vspace{-0.2cm}
	\caption{\small Image prediction error comparison between the reconstructed image from the predicted latent vectors over the number of time steps into the future and ground truth image observations, for test dataset of scenarios containing 2, 4, and 8 objects (plus the pusher). 2 and 8 objects is generalization over number of objects, 4 is as during training. One step corresponds to 2 cm movement, i.e.\ for 50 steps the pusher has moved 1 m.\vspace{-0.5cm}}
	\label{fig:baselinesImage}
\end{figure}

\begin{figure}
	\begin{minipage}[t]{9.8cm}
		\centering
		\captionsetup[subfloat]{captionskip=2pt}
		\vspace{-0.28cm}
		\tikzset{external/export next=false}
		\begin{tikzpicture}
			\node[] at (0,0) (a) {$t=0$};
			\node[] at (6,0) (b) {$t=38$};
			\draw[->] (a) -- (b);
		\end{tikzpicture}\\[-0.1cm]
		\subfloat[Predictions with our method]{
			\includegraphics[trim={1cm 1.2cm 1cm 0}, clip, width=1.1cm]{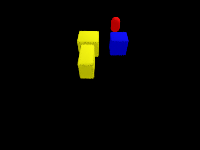}
			\includegraphics[trim={1cm 1.2cm 1cm 0}, clip,width=1.1cm]{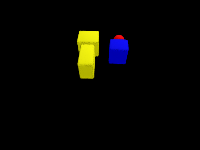}
			\includegraphics[trim={1cm 1.2cm 1cm 0}, clip,width=1.1cm]{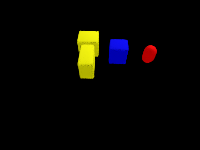}
			\includegraphics[trim={1cm 1.2cm 1cm 0}, clip,width=1.1cm]{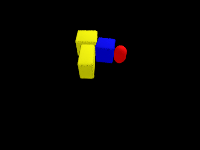}
			\includegraphics[trim={1cm 1.2cm 1cm 0}, clip,width=1.1cm]{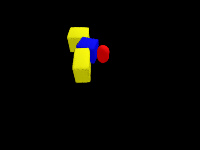}
			\includegraphics[trim={1cm 1.2cm 1cm 0}, clip,width=1.1cm]{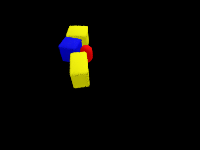}
			\label{fig:forwardPred3Objects:ours}
		}\\[-0.01cm]
		\subfloat[Predictions with dense adjacency matrix baseline]{
			\includegraphics[trim={1cm 1.2cm 1cm 0}, clip,width=1.1cm]{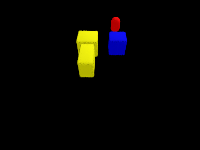}
			\includegraphics[trim={1cm 1.2cm 1cm 0}, clip,width=1.1cm]{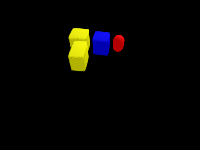}
			\includegraphics[trim={1cm 1.2cm 1cm 0}, clip,width=1.1cm]{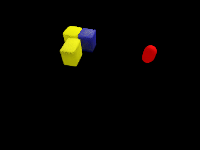}
			\includegraphics[trim={1cm 1.2cm 1cm 0}, clip,width=1.1cm]{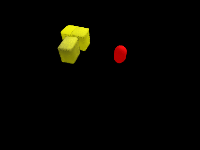}
			\includegraphics[trim={1cm 1.2cm 1cm 0}, clip,width=1.1cm]{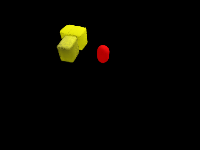}
			\includegraphics[trim={1cm 1.2cm 1cm 0}, clip,width=1.1cm]{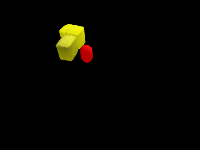}
			\label{fig:forwardPred3Objects:denseA}
		}
		\\[-0.01cm]
		\subfloat[Predictions with CNN decoder baseline (no NeRF)]{
			\includegraphics[trim={1cm 1.2cm 1cm 0}, clip,width=1.1cm]{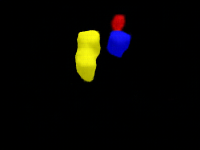}
			\includegraphics[trim={1cm 1.2cm 1cm 0}, clip,width=1.1cm]{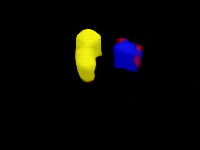}
			\includegraphics[trim={1cm 1.2cm 1cm 0}, clip,width=1.1cm]{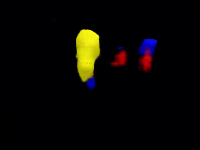}
			\includegraphics[trim={1cm 1.2cm 1cm 0}, clip,width=1.1cm]{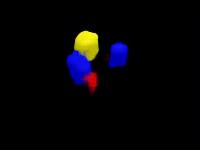}
			\includegraphics[trim={1cm 1.2cm 1cm 0}, clip,width=1.1cm]{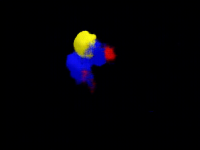}
			\includegraphics[trim={1cm 1.2cm 1cm 0}, clip,width=1.1cm]{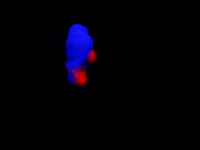}
			\label{fig:forwardPred3Objects:CNNDecoder}
		}	
		\vspace{-0.2cm}	
		\caption{\small Visual forward predictions. With our proposed method (a), the predictions are very sharp, even after 38 steps, while with a dense adjacency matrix (b) leads to drifting objects until the predictions are not useful anymore. The CNN decoder baseline is even worse, such that after only a few steps the predictions are of little use.  Multiple object interactions happen in this scene.
		}
		\label{fig:forwardPred3Objects}
	\end{minipage}%
	\hfill
	\begin{minipage}[t]{3.5cm}
		\captionsetup[subfloat]{captionskip=2pt}
		\subfloat[Ground truth]{
			\includegraphics[width=1.7cm]{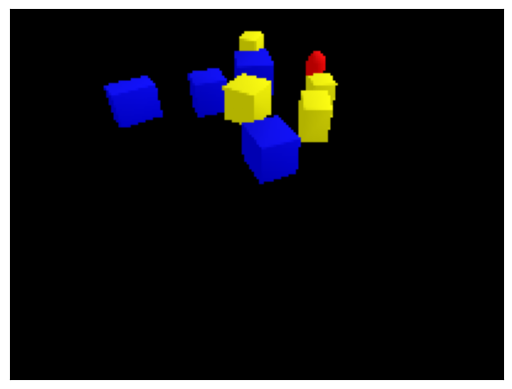}
			\includegraphics[width=1.7cm]{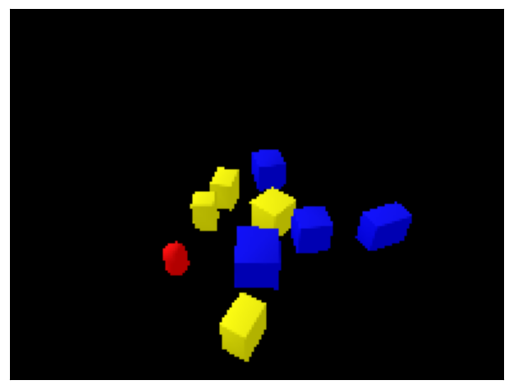}
		}\\[-0.1cm]
		\subfloat[Reconstruction]{
			\includegraphics[width=1.7cm]{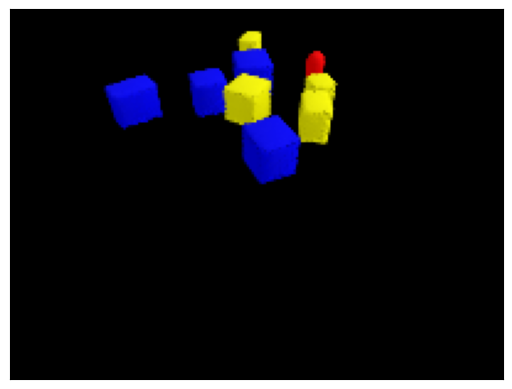}
			\includegraphics[width=1.7cm]{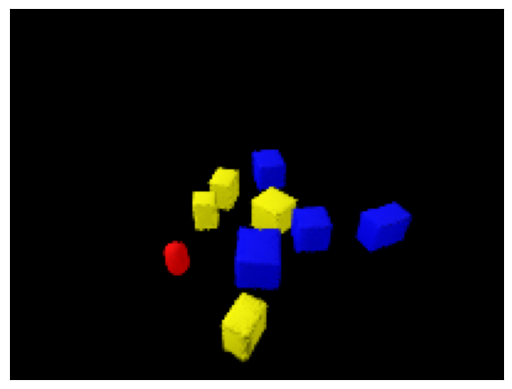}
		}
		\caption{\small Generalization to twice as many objects as during training.}
		\label{fig:eightObjects}
	\end{minipage}
	\vspace{-0.6cm}
\end{figure}

\textbf{Reconstruction and Prediction Performance for Generalization over Numbers of Objects.}
Fig.~\ref{fig:forwardPred3Objects:ours} shows predictions of the model forward unrolled in time for an action sequence of the red pusher, i.e.\ applying Algo.~\ref{algo:forwardPred} (appendix) to an initial scene observation and rendering the predicted latent vectors with the NeRF decoder.
Despite the movements in this scene leading to multiple object interactions, even after 38 time steps, the rendered predictions from the model are still sharp and reflect the underlying dynamics.
By utilizing the estimated adjacency matrix, there is little drift in the objects, leading to long-term prediction stability.
Due to its compositional nature, our model generalizes to scenes that contain more or less objects than in the training set, as shown in Fig.~\ref{fig:eightObjects} where eight objects plus the pusher are observed and reconstructed with high quality from novel views, although during training the model has seen only and exactly 4 objects.

\textbf{Comparison to Non-Compositional Scene Representation Baselines.}
We compare to two non-compositional baselines where the scene is represented globally with one single latent vector per time-step.
The dynamics model for these baselines is an MLP $z^{t+1} = F_\text{MLP}(z^t, q)$ that takes the action $q$ as an additional input.
The first baseline (Global NeRF) is the approach from \cite{li20223d}, i.e.\ we use their CNN encoder instead of our implicit object encoder producing one latent vector conditioning a global NeRF that reconstructs the whole scene.
The second baseline (Global 2D CNN auto encoder) uses both a 2D CNN encoder and 2D CNN decoder as well as a single latent vector representing the whole scene. 
Such frameworks have been used many times in the literature, e.g.\ \cite{ebert2018visual,watter2015embed,hafner2019learning,hafner2019dream,schrittwieser2020mastering} and are known as visual foresight.
Fig.~\ref{fig:baselinesImage} shows that both global baselines are significantly inferior in our scenarios to our proposed compositional framework, especially for long horizons.

\textbf{Comparison to 2D Baselines -- Importance of NeRF as Decoder.}
In this section, we replace the NeRF decoder with a 2D CNN decoder to investigate the importance of NeRFs.
This decoder takes as input one single latent vector and the camera matrix, i.e.\ $I = D_\text{CNN}(z, K)$.
In order to make it compositional, we aggregate the set of latent vectors $z_{1:m}$ from $\Omega$ with a mean operation and then pass the aggregated feature through an MLP to produce the single $z$ for $D_\text{CNN}$.
The rest of the architecture, i.e.\ implicit object encoder and GNN, stays the same.
Since there is no clear way to estimate the adjacency matrix from $D_\text{CNN}$, we use a dense adjacency matrix for the GNN.
As one can see in Fig.~\ref{fig:baselinesImage}, the long-term prediction performance of the CNN decoder is significantly worse than with a compositional NeRF model as the decoder, especially when asking for numbers of objects that differ from the training distribution.
Qualitatively, one can see in Fig.~\ref{fig:forwardPred3Objects:CNNDecoder} that not only the initial reconstruction is much less sharp compared to the NeRF-based models, but especially also that even after only a few time-steps, the predictions with the CNN decoder are of little use.

\textbf{Comparison to CNN Encoder.}
Exchanging the implicit object encoder with a 2D CNN compositional encoder leads to an auto-encoder framework similar to \cite{stelzner2021decomposing}.
As seen in Fig.~\ref{fig:baselinesImage}, the performance is better compared to the other baselines, but still clearly worse than with the proposed method.

\textbf{Importance of Estimating the Adjacency Matrix.}
In Sec.~\ref{sec:GNN}, we propose how the adjacency matrix of the GNN can be estimated from the learned NeRFs to increase the long-term stability of the predictions.
Here we compare to a dense adjacency matrix, i.e.\ where the network has to figure out from the latent vectors themselves which objects interact.
As one can see in Fig.~\ref{fig:baselinesImage} and Fig.~\ref{fig:forwardPred3Objects:denseA}, a dense $A$ has significantly worse long-horizon prediction performance compared to our proposed way of estimating $A$ through the learned NeRF model.
In the 2 and 8 object case (generalization over numbers of objects), the predictions with the dense $A$ are useless after only a few time-steps.

\textbf{Non-Compositional Dynamics.}
Replacing the GNN with a fully connected MLP $z_{1:m}^{t+1} = F_\text{MLP}(z_{1:m}^t)$ leads to worse performance than with a GNN with dense adjacency matrix.
This model cannot generalize to different numbers of objects due to its fixed input size.

\textbf{Summary of Performance Comparisons}
Our method outperforms all baselines both in terms of pure reconstruction error (as can be seen in Fig.~\ref{fig:baselinesImage} by the error after 0 prediction steps) \emph{and} its ability to perform long-term predictions forward unrolled on the model's own predictions.
Estimating the adjacency matrix from the model itself is important for long-term stability as it prevents objects from drifting away.
Too large drift makes future predictions for a pushing tasks meaningless.
Since the reconstruction error of our proposed method without dynamics is better than the baselines, the question arises if the increased performance is an artifact of the lower reconstruction error.
We show in Fig.~\ref{fig:baselinesPredRecon} the error in the image space between renderings when having access to the observations at each step and the renderings from the predicted latent vectors into the future after observing the scene only at the beginning.
This shows the increase in error relative to the reconstruction process.
The results indicate that not solely the reconstruction itself is the reason for the better performance, but that the structural choices of our framework also enable to learn the dynamics more precisely.

\vspace{-0.1cm}
\subsection{Planning and Execution Results on Object Sorting Task}
\begin{figure}
	\centering
	\vspace{-0.5cm}
	\captionsetup[subfloat]{captionskip=2pt}
	%\captionsetup[subfloat]{width=3cm}
	\subfloat[Initial scene]{
		\includegraphics[trim={2cm 1cm 2cm 2cm}, clip, width=2.2cm]{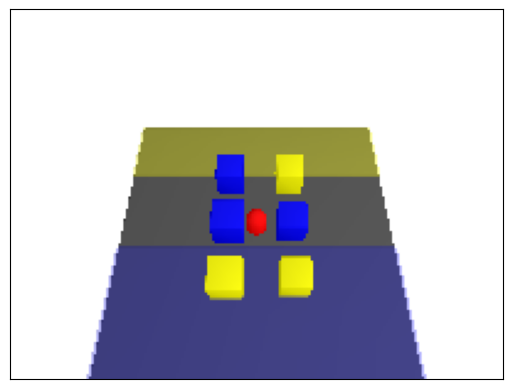}
		\label{fig:boxPlanning:initialScene6}
	}
	\subfloat[Final goal achieved]{
		\quad
		\includegraphics[trim={2cm 1cm 2cm 2cm}, clip, width=2.2cm]{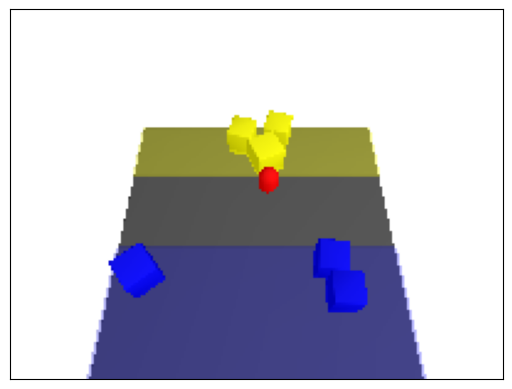}
		\quad
		\label{fig:boxPlanning:finalScene6}
	}
	\subfloat[Initial scene]{
		\quad
		\includegraphics[trim={3cm 1.5cm 0.3cm 0.5cm}, clip, width=2.2cm]{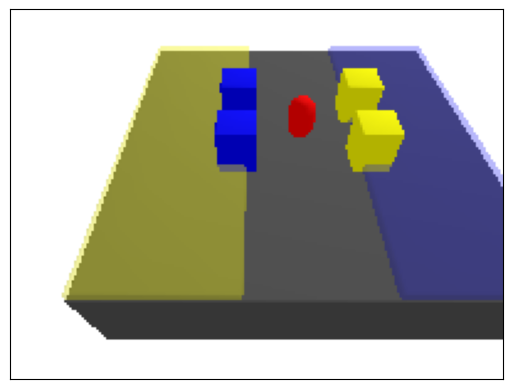}
		\quad
		\label{fig:boxPlanning:initialScene4}
	}
	\subfloat[Final goal achieved]{
		\quad
		\includegraphics[trim={3cm 1.5cm 0.3cm 0.5cm}, clip, width=3cm, width=2.2cm]{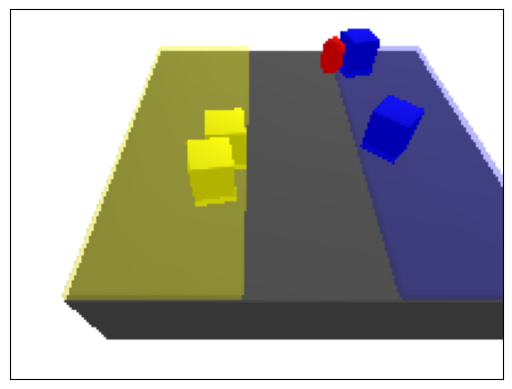}
		\quad
		\label{fig:boxPlanning:finalScene4}
	}
	\vspace{-0.15cm}
	\caption{\small Two planning scenarios with our learned visual dynamics model and latent space RRT. The goal is to move the blue and yellow boxes into their respective shaded areas. (a), (c) are initial states, (b), (d) shows the achieved goal at the end of the planning/execution loop.}
	\vspace{-0.1cm}
	\label{fig:boxPlanning}
\end{figure}

\vspace{-0.2cm}
To demonstrate the effectiveness of the learned model, we utilize it to solve a  box sorting task, where the pusher needs to push colored boxes into their corresponding goal regions as shown in Fig.~\ref{fig:boxPlanning}.
This task is inspired by \cite{pmlr-v164-florence22a} and involves multiple challenges: As multiple objects interact, a greedy strategy of pushing objects straight to the goal region fails.
Movements, i.e.\ actions, of the pusher do often not immediately lead to a change in the cost function, since contact with the object from a suitable side has to be established \cite{driess2022CoRL, schubert2021learning}.
In the appendix Sec.~\ref{app:planningRRT} we propose a latent space RRT that uses our framework for planning.
Refer to the appendix and the video for more details about our proposed planning algorithm and comparisons to baselines.

\vspace{-0.1cm}
\subsection{Real World Experiments}
\vspace{-0.2cm}
Fig.~\ref{fig:firstPage:pred} shows the rendered forward predictions of our model for a real world scenario where a robot pushes a shoe and a giraffe-shaped toy. 
Fig.~\ref{fig:firstPage:novel} are renderings from novel view points.

\vspace{-0.1cm}
\subsection{Applicability to Deformable Objects}\label{sec:exp:deformable}
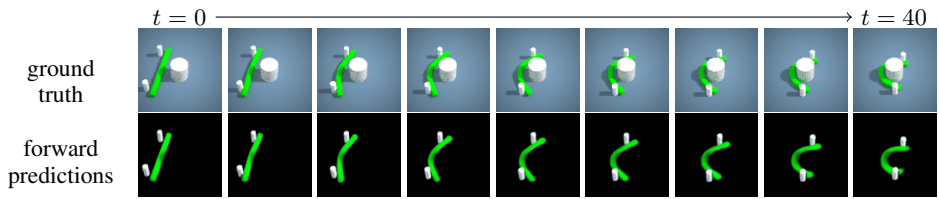
\begin{figure}
	\vspace{-0.2cm}
	\centering
	\tikzset{external/export next=false}
	\begin{tikzpicture}
	\node[] at (0,0) (a) {$t=0$};
	\node[] at (9.5,0) (b) {$t=40$};
	\draw[->] (a) -- (b);
	\node[] at (-2.4,0) {};
	\end{tikzpicture}\\[-0.1cm]
	\tikzset{external/export next=false}
	\begin{tikzpicture}
		\node[align=center, minimum width=1.9cm] {ground\\truth};
	\end{tikzpicture}
	\includegraphics[width=1.1cm]{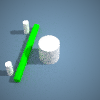}
	\includegraphics[width=1.1cm]{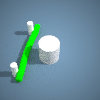}
	\includegraphics[width=1.1cm]{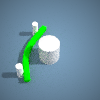}
	\includegraphics[width=1.1cm]{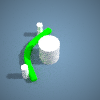}
	\includegraphics[width=1.1cm]{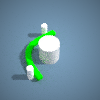}
	\includegraphics[width=1.1cm]{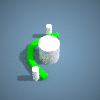}
	\includegraphics[width=1.1cm]{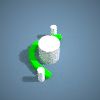}
	\includegraphics[width=1.1cm]{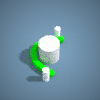}
	\includegraphics[width=1.1cm]{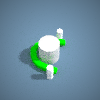}\\
	\tikzset{external/export next=false}
	\begin{tikzpicture}
	\node[align=center, minimum width=1.9cm] {forward\\predictions};
	\end{tikzpicture}
	\includegraphics[width=1.1cm]{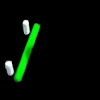}
	\includegraphics[width=1.1cm]{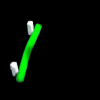}
	\includegraphics[width=1.1cm]{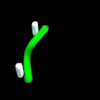}
	\includegraphics[width=1.1cm]{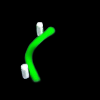}
	\includegraphics[width=1.1cm]{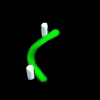}
	\includegraphics[width=1.1cm]{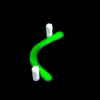}
	\includegraphics[width=1.1cm]{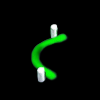}
	\includegraphics[width=1.1cm]{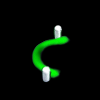}
	\includegraphics[width=1.1cm]{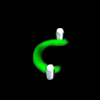}
	\caption{\small Forward predictions for deformable object scenario. The predicted reconstructions are based on the latent vector from the initial observation ($t=0$) that is then forward predicted with the dynamics model.}
	\label{fig:plasticine}
	\vspace{-0.5cm}
\end{figure}

\vspace{-0.2cm}
The experiments so far focused on objects that behave mainly like rigid objects when being pushed.
In Fig.~\ref{fig:plasticine} we show that our method is also applicable to deformable objects simulated with \cite{huang2021plasticinelab}.

\vspace{-0.1cm}
\section{Discussion \& Limitations}
\vspace{-0.3cm}
\textbf{Computational Efficiency.}
Our framework is computationally more demanding during inference time than 2D CNN decoder baselines, mainly due to NeRF evaluations.
Many methods have been developed to increase the speed of NeRF \cite{Reiser2021ICCV}, from which our framework could benefit.

\textbf{Object Masks.}
The compositional scene encoding framework requires object masks to achieve compositionality.
Many mature methods for instance segmentation have been developed such that we believe having masks is a reasonable assumption.
However, one could add an output to our implicit encoder that provides object labels or use mechanisms similar to slot attention \cite{stelzner2021decomposing}.

\textbf{Latent Representations.}
We have shown the great benefits of a compositional latent representation as it not only provides generalization over different numbers of objects in the scene, but also leads to increased reconstruction and dynamics prediction performance compared to non-compositional baselines.
Furthermore, latent representations compress observations, enabling efficient dynamics prediction.
However, as still each object in the scene is represented as a latent vector of finite size, latent models are capable of mainly representing objects with shapes similar to the training distribution.
To address this, compositionality could not only be introduced on the scene level, but also by representing objects themselves in a composable way.

\textbf{Long-Term Prediction Stability.}
Our dynamics model framework exhibits significantly better long-term prediction stability compared to baselines.
Our experiments indicate that this is due the structural biases enabled through (compositional) NeRFs. 
This stability allowed us to use the model for planning scenarios requiring long-horizons, which none of the baseline methods could support. 
However, we believe that there is still room for improvement regarding the prediction stability.
Especially for deformable objects, we observed that after many prediction steps objects in the scene are predicted to penetrate or move through each other, which could be improved in future work.

\vspace{-0.1cm}
\section{Conclusion}
\vspace{-0.3cm}
Visual dynamics models are of high interest to the computer vision and robotics community, as they avoid explicit shape model assumptions and imply end-to-end perception.
However, to support manipulation planning and reasoning, we need models that generalize strongly over objects and provide stable long-term predictions.
In this paper we proposed a system that introduces 3D structural and compositional priors at various levels, namely compositional NeRFs, 3D implicit object encoders, and GNNs dynamics with an adaptive adjacency matrix.
Together our system exhibits significantly stronger long-term prediction performance compared to multiple baselines without these priors or without compositionality, and supports using a latent space RRT planner.
We have shown generalization over different numbers of objects, notably up to two times more than during training.

\clearpage

\acknowledgments{
This research has been supported by the Deutsche Forschungsgemeinschaft (DFG, German Research Foundation) under Germany’s Excellence Strategy – EXC 2002/1 “Science of Intelligence” – project number 390523135.
Danny Driess thanks the International Max-Planck Research School for Intelligent Systems (IMPRS-IS) for the support.
The authors thank Valentin Hartmann for discussions regarding RRTs.}

%\bibliography{references}

\clearpage

\appendix

\section{Expanded Related Work}\label{app:relatedWork}

\subsection{Model-Based Planning in Robotic Manipulation}

Model-based planning algorithms typically build a dynamics model of the environment and then use the model to plan the agent's behavior in order to minimize some task objectives.
We can roughly categorize the methods by whether the model is constructed from first principles (i.e., physical rules) or learned from data (i.e., data-driven models).
Physics-based models typically require complete information about the objects' geometry and the system's state~\cite{hogan2016feedback,zhou2019pushing,driess2020deep,20-toussaint-RAL,hartmann2021long}, which limits their applicability in robotic manipulation tasks involving unknown object models and partially observable states.
Data-driven methods, on the other hand, learn a dynamics model directly from the robot's interaction with the environment and have shown impressive results in manipulation tasks ranging from closed-loop planar pushing~\cite{bauza2018data} to complicated dexterous manipulation~\cite{nagabandi2020deep}.
Many of the data-driven planning frameworks learn dynamics models directly from visual observation based on representations defined at different levels of abstraction, such as pixel space~\cite{finn2016unsupervised,ebert2017self,schenck2018perceiving,ebert2018visual,suh2020surprising,driess2021icra}, 3D volumetric space~\cite{xu2020learning}, signed-distance fields~\cite{driess2022CoRL, strecke2021_diffsdfsim, wi2022virdo}, keypoint space~\cite{kulkarni2019unsupervised,manuelli2020keypoints,li2020causal}, and low-dimensional latent space~\cite{watter2015embed,hafner2019learning,hafner2019dream,schrittwieser2020mastering}.
Approaches commonly employ an image reconstruction loss~\cite{hafner2019learning,hafner2019dream}, an self-supervised time contrastive loss~\cite{sermanet2018time}, or jointly train a forward and an inverse dynamics model~\cite{agrawal2016learning} to make sure that the representation encodes meaningful information about the environment.
Our method takes a step forward by learning graph-based latent representations from visual observations. The learned model accurately encodes the underlying 3D contents, allowing our learned model to achieve precise manipulation of compositional environments and generalize outside the training distribution, i.e. to scenes with more (and less) objects than during training.

\clearpage

\section{Details -- Encoding Scenes with Compositional Image-Conditioned NeRFs}\label{app:autoencoder}

This section provides details -- especially visualizations of the network architectures -- on our proposed compositional auto-encoder framework.

\begin{figure}[!b]
	\centering
	\scalebox{0.7}{\input{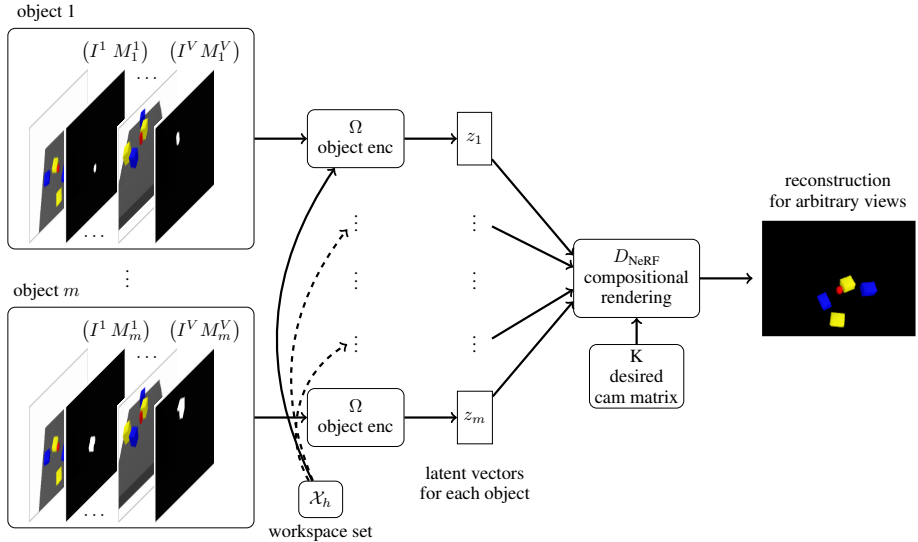}}
	\caption{Overview of the image-conditioned compositional NeRF autoencoder structure. Each object $j=1,\ldots, m$ is observed in terms of RGB images $I^i$ of the scene from multiple views $i=1,\ldots, V$ with corresponding object masks $M_j^i$ and camera matrices $K^i$. These inputs are encoded with the implicit object encoder $\Omega$ for each object separately, yielding latent vectors $z_j$ for each object $j$ (Sec.~\ref{sec:implicitObjectEncoder}). The weights of the object encoder $\Omega$ are shared between all objects. Further, the workspace set $\mathcal{X}_h$ where the implicit object feature functions $y_j(\cdot)$ within the encoder $\Omega$ are queried are the same for all objects. Refer to Fig.~\ref{fig:omega} and Fig.~\ref{fig:omega:Phi} for further details on the architecture of $\Omega$. The set of latent vectors of all objects $z_{1:m}$ is rendered into an image for an arbitrary view specified by $K$ via compositional NeRF rendering (Sec.~\ref{sec:decoder}).}
	\label{fig:objectEncoderOverview}
\end{figure}

\subsection{Encoder}
Fig.~\ref{fig:objectEncoderOverview} visualizes the whole auto-encoder architecture, where the encoder
\begin{align}
	z_{1:m} = \Omega\left(I^{1:V}, K^{1:V}, M^{1:V}_{1:m}, \mathcal{X}_h\right)
\end{align}
maps posed images $(I, K)^{1:V}$ from $V$-many views of $m$-many objects including their object masks $M^{1:V}_{1:m}$ as well as an workspace set $\mathcal{X}_h$ to a set of latent vectors $z_{1:m}$ describing the objects in the scene.
The same object encoder $\Omega$ and workspace set $\mathcal{X}_h$ is used for all objects.
In particular, $\mathcal{X}_h$ is \emph{not} a 3D bounding box for an individual object, but covers the whole workspace of the scene.
See Fig.~\ref{fig:sceneBB} for a visualization of the workspace set $\mathcal{X}_h$.

\begin{figure*}
	\centering
	\subfloat[Feature encoder $E$ for view $i$ of image-conditioned implicit object encoder. $E$ first projects the query point $x\in\mathbb{R}^3$ into the camera coordinate system via $K^i$. This projected coordinate is, on the one hand, encoded into a feature via an MLP. On the other hand, the pixel coordinates $\big(u^i(x), v^i(x)\big)$ are used to obtain the RGB value at the projected $x$ via bilinear interpolation. The coordinate feature and the interpolated RGB value are concatinated and further processed with an other MLP, leading to the final output $E\left(I^i, K^i(x)\right)\in\mathbb{R}^{n_o}$.]{
		\begin{minipage}{\textwidth}
			\centering
			\scalebox{0.9}{\input{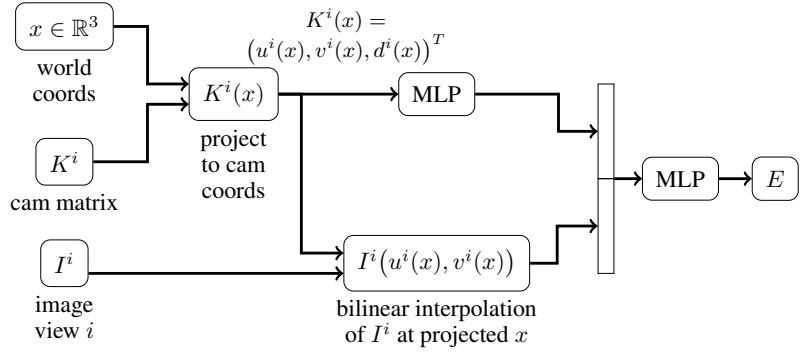}}
		\end{minipage}
		\label{fig:omega:E}
	}\\
	\subfloat[Implicit object feature function $y_j$. The point $x\in\mathbb{R}^3$ in world coordinates where $y_j$ is queried is projected into the camera coordinate systems of each view $I^i$ where the (shared) feature encoder $E$ (Fig.~\ref{fig:omega:E}) produces a feature of the object $j$ at the projected $x$. The masks $M^i_j$ determine whether the computed feature is relevant for aggregating the features from the different views into the final feature for object $j$. This aggregation is denoted with the sum symbol $\Sigma$, refer also to \eqref{eq:yObjectFunction}.]{
		\begin{minipage}{\textwidth}
			\centering
			\scalebox{0.9}{\input{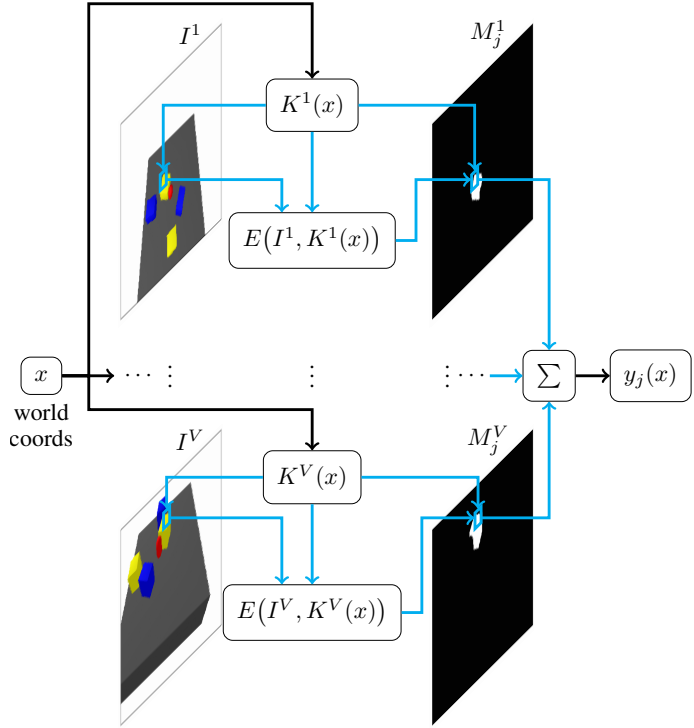}}
		\end{minipage}
		\label{fig:omega:y}
	}
	\caption{Visualization of the internals of the implicit object encoder $\Omega$ as a function of the RGB image observations $I^{1:V}$, their camera matrices $K^{1:V}$, and object masks $M^{1:V}_j$ of the object $j$ for views $i=1,\ldots, V$. (a) shows the architecture of the feature encoder $E$. (b) visualizes how the implicit object feature function $y$ \eqref{eq:yObjectFunction} aggregates the features produced by $E$ queried on the different views into the feature at the query point $x\in\mathbb{R}^3$ (world coordinates) by taking into account the object masks $M^i_j$ in the different views.  See Fig.~\ref{fig:omega:Phi} for the volumetric object encoder $\Phi$ that turns $y_j$ into the latent vector $z_j\in\mathbb{R}^k$.}
	\label{fig:omega}
\end{figure*}

\begin{figure}
	\centering
	\scalebox{0.9}{\input{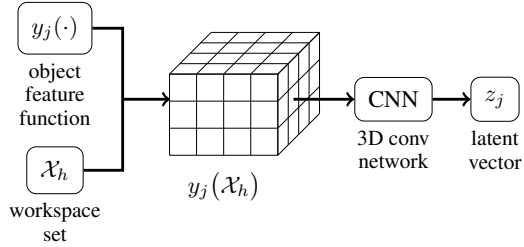}}
	\caption{Volumetric object encoder $\Phi$. The object feature function $y_j$ of object $j$ is evaluated on the discretized workspace set $\mathcal{X}_h$ which produces an object feature voxel grid $y_j\big(\mathcal{X}_h\big)$ that is turned into the latent vector $z_j\in\mathbb{R}^k$ via a 3D convolutional network.}
	\label{fig:omega:Phi}
\end{figure}

Internally, $\Omega$ consists of a feature encoder $E$ that outputs an $n_o$-dimensional feature vector from the image $I^i$ of view $i$ at any 3D world coordinate $x\in\mathbb{R}^3$
\begin{align}
	E(I^i, K^i(x)) \in\mathbb{R}^{n_o}.
\end{align}
Similar architectures of computing such pixel features from world coordinates have been proposed, e.g., in \cite{yu2021pixelnerf, saito2019pifu, ha2021learning} for the \emph{single} object case.
However, we use $E$ quite differently compared to these works, as we compute a latent vector from pixel-aligned features with 3D convolutions.
The object feature function $y_j$, cf.\ \eqref{eq:yObjectFunction}, aggregates the features from the individual views by taking into account the masks of object $j$ in each view.  
The architectures of $E$ and $y$ are visualized in Fig.~\ref{fig:omega}.
Fig.~\ref{fig:omega:Phi} shows how an object feature function $y_j(\cdot)$ for object $j$ is turned by $\Phi$ into its corresponding latent vector $z_j$ by querying it on the workspace set $\mathcal{X}_h$ followed by a 3D convolutional 
network.

\newpage
\subsection{Background on Neural Radiance Fields}\label{app:backgroundNerf}
This section summarizes non-conditional, global Neural Radiance Fields (NeRFs) for the purposes of this work in more detail as in the main paper.
Refer to Sec.~\ref{sec:decoder} on how we build a compositional decoder from NeRFs.
For even further details on vanilla NeRFs we refer to the original publication \cite{mildenhall2020nerf}.
The general idea of NeRF is to learn a function $f$ that predicts, at a 3D world coordinate $x\in\mathbb{R}^3$, the (emitted) RGB color value $c(x)\in\mathbb{R}^{3}$ and volume density $\sigma(x)\in\mathbb{R}_{\ge0}$.
Based on the learned $(\sigma(\cdot), c(\cdot)) = f(\cdot)$, an image from an arbitrary view and camera configuration can be rendered by determining the color $C(r)\in\mathbb{R}^3$ of each pixel along its corresponding camera ray $r(\alpha) = r(0) + \alpha d$ through
\begin{align}
C(r) = \int_{\alpha_n}^{\alpha_f}T_f(r, \alpha)\sigma(r(\alpha))c(r(\alpha))\diff \alpha \label{eq:nerfC}
\end{align}
with
\begin{align}
T_f(r, \alpha) = \exp\left(-\int_{\alpha_n}^\alpha \sigma(r(s))\diff s \right).\label{eq:nerfT}
\end{align}
Here, $r(0)\in\mathbb{R}^3$ is the camera origin, $d\in\mathbb{R}^3$ the pixel dependent direction and $\alpha_n, \alpha_f \in \mathbb{R}$ the near and far bounds within an object is expected, respectively. 
The function $f$ is a fully-connected neural network and the integrals in \eqref{eq:nerfC} and \eqref{eq:nerfT} are estimated by a simple quadrature rule, see \cite{mildenhall2020nerf}, which make the whole rendering process differentiable and hence trainable with stochastic gradient descent.
In most NeRF formulations, $f$ takes a view direction as an additional input, which is beneficial to reconstruct reflections and other lighting effects.
For the scenario we consider in this work, we found that incorporating view directions was not necessary and therefore omitted them.
Including a view direction is a straightforward extension to what we present here.

\subsection{Training}\label{app:encoderDecoderTraining}

The auto-encoder framework is trained end-to-end on an $\text{L}_2$ image reconstruction loss.
Since, as mentioned, solely the objects are represented as NeRFs and not the background, we compute the union of the masks of the individual objects
\begin{align}
M_\text{tot}^i = \bigvee_{j=1}^m M_j^i
\end{align}
and define the target image in a view as $I^i\circ M_\text{tot}^i$ with $\circ$ denoting the element-wise product.

A known issue of NeRF is its computational efficiency \cite{stelzner2021decomposing}, since for every pixel all $f_j$'s have to be queried on many points along the camera ray.
We make two simple, but important, improvements to reduce the computational demand.

First, the near and far bounds $\alpha_n$, $\alpha_f$ are determined individually for each camera ray such that only those points along the rays that are within the workspace set $\mathcal{X}$ are considered. 
This is a reasonable assumption since we assumed that the objects are in the workspace set in the first place.
That way, the computational efficiency is already greatly increased by reducing the number of points where functions $f_j$'s have to be queried. 

\begin{figure*}
	\centering
	\subfloat[Scene observation $I^i$]{
		\includegraphics[width=4.cm]{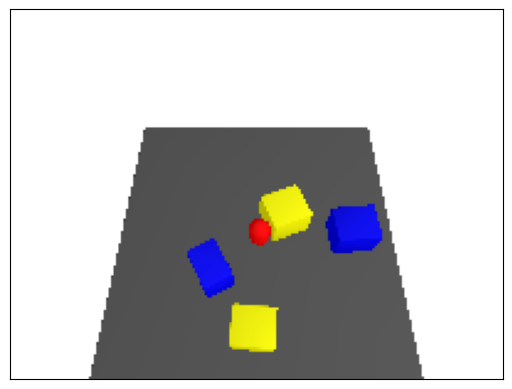}
	}\hspace{0.5cm}
	\subfloat[Training target $I^i\circ\hat{M}_\text{tot}^i$ with enlarged mask. White area: no rays are considered]{
		\includegraphics[width=4.cm]{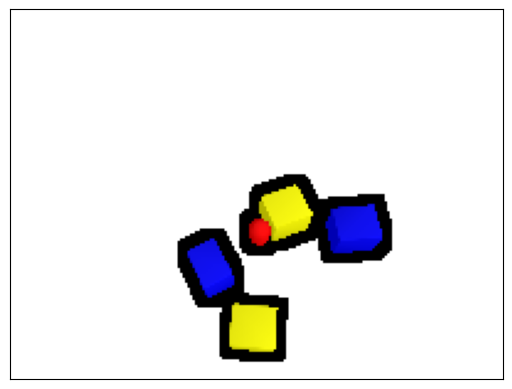}
		\label{fig:trainingMasks:trainingMasks}
	}
	\hspace{0.5cm}
	\subfloat[Prediction by learned model rendered everywhere and not only at the masks]{
		\includegraphics[width=4.cm]{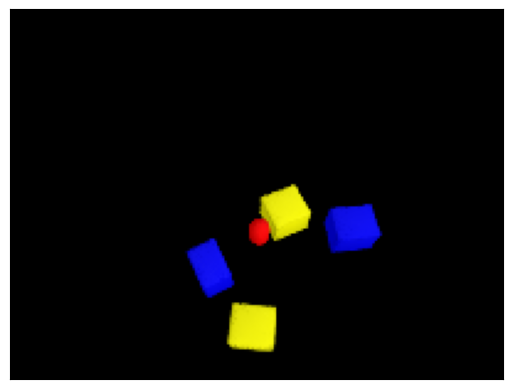}
	}
	\caption{Visualization of the training target for an example scene in the training dataset. Same scene as in Fig.~\ref{fig:sceneBB}.}
	\label{fig:trainingMasks}
\end{figure*}

Moreover, as the scenes we consider in our experiments (see for example Fig.~\ref{fig:sceneBB}) are composed of multiple smaller objects, when masking out the background, the majority of pixels in each view is black and therefore does not contain information about the scene, although the model is evaluated on those areas.
To further decrease the number of points where the NeRFs have to be queried, we only consider those rays for a view that pass through the mask of at least one object in that view.
It turned out, however, that training only on rays that go through $M_\text{tot}$ leads to blurry reconstructions, since there is no loss indicating that the objects should end outside of the masks.
In order to resolve this, we enlarge the combined mask $M_\text{tot}^i$ of a view with a convolution operation by a few pixels.
We denote this enlarged mask by $\hat{M}_\text{tot}^i$.
Together, these techniques ensure that the model learns sharp object boundaries, while significantly reducing the number of considered rays and required NeRF evaluations.
See Fig.~\ref{fig:trainingMasks} for a visualization of this procedure.

These considerations lead to the following training objective of $D_\text{NeRF}$ and $\Omega$ for a view $i$
\begin{align}
&\mathcal{L}^i = \sum_{(u,v)\in\hat{M}_\text{tot}^i}\notag \\
&~\left\|\left(I^i\circ M_\text{tot}^i\right)_{uv} \!\!\!-\! D_\text{NeRF}\!\left(\Omega\left(I^{1:V}, K^{1:V}, M^{1:V}_{1:m}, \mathcal{X}_h\right), K^i\right)_{uv}\right\|_2^2.
\end{align}
During training, we randomly sample a view from the dataset for each mini-batch and update the parameters of $D_\text{NeRF}$ and $\Omega$ using the ADAM optimizer \cite{kingma2014adam}.

Another side effect of training on the enlarged masks $\hat{M}_\text{tot}^i$ only is that it improved the training stability and reconstruction qualities of the model.
Indeed, when we trained the model on the whole image, depending on the weight initialization of the network, the model sometimes very quickly converged to a state where it only predicted a black image, since the majority of pixels are actually black and hence a low loss could be achieved.
Training on the enlarged masks prevents this reliably.

\subsection{Rigid Transformations and Novel Scene Generation}\label{app:novelSceneGeneration}

The compositional formulation of our model makes it trivial to add and remove objects from the scene.
Furthermore, since the proposed object representation $y_j$ is a function of a 3D coordinate, we can rigidly transform objects in the workspace by applying a rigid transformation \cite{driess2022CoRL}.
Let $R(q)\in\mathbb{R}^{3\times 3}$ and $s(q)\in\mathbb{R}^3$ be a rotation matrix and translation vector as a function of $q\in\mathbb{R}^7$ (translation + quaternion), respectively.
Then,
$
y_j\left(R(q)^T(~\cdot~ - s(q))\right)
$
is the object feature function transformed by $q$.
Consequently, evaluating the transformed $y_j$ on $\mathcal{X}_h$ with $\Phi$ from \eqref{eq:Phi} leads to a new latent vector that represents the object $j$ being transformed by $q$, which we denote with
\begin{align}
\!\mathcal{T}(q)[z_{j}] = \Omega\left(I^{1:V}, K^{1:V}, M^{1:V}_{j}, R(q)^T\left( \mathcal{X}_h-s(q)\right)\right).\! \label{eq:zRigidTransormation}
\end{align}
Please note the slight abuse of notation here, the term $R(q)^T\left( \mathcal{X}_h-s(q)\right)$ has to be understood elementwise for each entry in $\mathcal{X}_h$.

Composing scenes via rigid transformations applied to the input of the individual NeRFs $f$ has been considered before, e.g.\ in \cite{Niemeyer2020GIRAFFE}.
However, transforming a NeRF by applying the rigid transformation to its $x$ input only leads to changes in the rendered visual space, i.e.\ it has, in particular, no influence on the latent vectors of the objects.
Since we want the latent vectors to represent not only the appearance of an individual object, but the geometric information of the object within the scene relative to other objects, just transforming the NeRF models is not sufficient.
Therefore, using \eqref{eq:zRigidTransormation} we get the latent vector rigidly transformed by $q$, which is crucial for our downstream dynamics prediction task.

This rigid transformation is important to define actions for the GNN dynamics model, as described in Sec.~\ref{app:actions} and Algo.~\ref{algo:forwardPred}.

\section{Details -- Graph Neural Network Latent Dyanmics Model}\label{app:gnn}
This section provides details about the latent dynamics model.
Especially relevant is Sec.~\ref{app:algoForwardPred} and Algo.~\ref{algo:forwardPred} where we describe the forward prediction algorithm.

\begin{figure*}
	\centering
	\scalebox{0.7}{\input{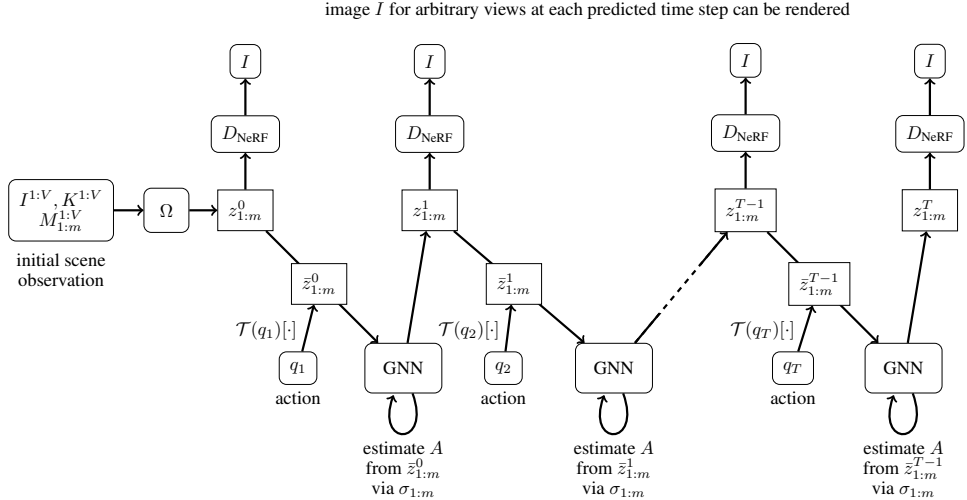}}
	\caption{Visualization of the forward prediction Algorithm \ref{algo:forwardPred}. After the initial scene observation is encoded into the latent vectors $z_{1:m}^0$, the GNN dynamic model is used to forward predict the evolution of the latent vectors in time for actions $q_t$ from the initial observation only. These actions rigidly transform the articulated objects (in the experiments the pusher), leading to $\bar{z}_{1:m}^t$ which is the input to the GNN to produce $z^{t+1}_{1:m}$. The adjacency matrix for the GNN dynamics model is estimated from the density prediction $\sigma(\cdot, z_j)$ for all objects from the predicted $z_j$ at each time step, see Sec.~\ref{app:adjacency}. The predicted latent vectors $z^t_{1:m}$ at time $t$ can be used to render images from arbitrary views or to reconstruct the scene.}
	\label{fig:forwardPredDiagram}
\end{figure*}

\subsection{Propagation Networks}\label{app:propNets}
Due to the compositional nature of the scenarios we consider, we require a dynamics model that maintains the capabilities of our auto-encoder to generalize over changing numbers of objects, for which graph neural networks (GNNs) are a natural choice.

The general idea behind learning dynamics models with GNNs is to associate each object in the scene with a node in a graph, which, in our case, means that each node in the graph is a latent vector $z_j$.
Edges between the nodes indicate if objects interact, e.g.\ by exchanging forces due to contact.
As argued in \cite{li2019propagation}, applying a simple GNN to the problem of dynamics prediction is problematic, since interactions caused at one node can influence not only the neighboring nodes, but higher-order neighbors.
For example, if three objects touch, the effects of applying a force at the first object have to propagate.
The scenarios we consider in the experiments contain multiple objects such that more than two objects can interact in one time step.
To take this into account, we use a message passing architecture inspired by \cite{li2019propagation}.

Let $z_i$ and $z_j$ be the latent vectors of objects $i$ and $j$. 
An edge encoder network $F_e$ determines a feature
\begin{align}
e_{ij} = F_e(z_i, z_j) \in \mathbb{R}^{n_e}\label{eq:edgeNetwork}
\end{align}
describing the interaction between the objects $i$, $j$.
An adjacency matrix $A\in\left\{0,1\right\}^{m\times m}$ has entry $A_{ij} = 1$ if object $i$ is influenced by object $j$.
Assume the state of all latent vectors $z_{1:m}^t$ at time $t$ is known.
The node propagator network $F_z$ recursively is queried $L$ many times to propagate the state  $z_{1:j}^t$ to the next time step $t+1$ as follows:
\begin{align}
l=1,\ldots, L ~:~~~\prescript{l}{}{z_i}^{t+1} = F_z\Big(z_i^t, \sum_{j ~:~ \prescript{l-1}{}{A}_{ij}^t = 1} \prescript{l-1}{}{e}_{ij}^{t}\Big)\label{eq:nodePropNetwork}
\end{align}
with $\prescript{0}{}{z_i}^{t+1} = z_i^{t}$, $\prescript{l}{}{e}_{ij}^{t} = F_e\left(\prescript{l}{}{z_i}^{t+1}, \prescript{l}{}{z_j}^{t+1}\right)$ for $l=0,\ldots,L-1$ and the final new predicted state $z^{t+1}_i = \prescript{L}{}{z}_i^{t+1}$.

\subsection{Adjacency Matrix from Learned Model}\label{app:adjacency}
The adjacency matrix $A$ in the GNN dynamics model \eqref{eq:nodePropNetwork} plays an important role in indicating which objects interact.
While a dense adjacency matrix, i.e.\ a graph where each node is connected to every other node implying that each object interacts with all other objects in the scene, would in principle work as the network could figure out from the latent representations itself which objects interact, we found that the long-horizon prediction performance is greatly increased if $A$ is more selective in reflecting which objects actually interact (refer to the experiments in Sec.~\ref{app:exp:adjacency}).
This is especially relevant for compositional scenes as considered in this work where there are many objects, but which often do not interact with each other in every timestep.

A central question is how the adjacency matrix can be obtained from the observations of the scene without manually specifying it.
Due to our model having strong 3D priors, we can exploit the density prediction $\sigma_j$ as defined in \eqref{eq:compNeRFs} for each object to determine the adjacency matrix from the models' own predictions during training and planning. 
In order to do so, for a threshold $\kappa \ge 0$, the collision integral
\begin{align}
S_{ij} = \int_{\mathcal{X}}[\sigma(x, z_i)>\kappa][\sigma(x, z_j)>\kappa]\diff x\label{eq:collisionIntegral}
\end{align}
over the density predictions of the learned NeRF model for objects $i$ and $j$ indicates if the two objects overlap or not.
A similar integral as in \eqref{eq:collisionIntegral} has been proposed in \cite{driess2022CoRL} to estimate collisions from signed-distance functions.
Based on this integral, we define the entries of the adjacency matrix between objects $i$ and $j$ as
\begin{align}
A_{ij} = \begin{cases}
1 & S_{ij} > 0\\
0 & \text{else}
\end{cases},
\end{align}
which implies that only those objects that are or are close to being in contact potentially interact.
Estimating $A$ this way takes the actual geometry of the objects in the scene into account.
In relation to the node propagation network $\eqref{eq:nodePropNetwork}$, this means that the adjacency matrix at step $l$ of the propagation becomes a function of the node encodings itself, i.e.\ 
\begin{align}
\prescript{l}{}{A}_{ij}^t = \prescript{l}{}{A}_{ij}^t\left(\prescript{l}{}{z_i}^{t+1}, \prescript{l}{}{z_j}^{t+1}\right).   
\end{align}
For training the GNN, however, changing the adjacency matrix during prediction is not differentiable. 
Therefore, we compute the adjacency matrices from the model such that they are constant within one time-step as follows.
We first compute an occupancy grid
\begin{align}
S_j = \left[\sigma(\mathcal{X}_h, z_j)>\kappa\right] \in\left\{0,1\right\}^{d\times h\times w}
\end{align}
for each object $j$ over the discretized workspace set $\mathcal{X}_h$ and then apply a 3D convolution operation on $S_j$ with a kernel consisting of only ones to expand the occupancy grid.
The now constant within one time-step $t$ entries $\prescript{l}{}{A}_{ij}^t = A_{ij}^t$ for all $l=0\ldots,L-1$ are then determined by checking if there is a voxel cell where both enlarged $S_i$ and $S_j$ have value one.
The size of the convolution kernel is chosen large enough such that the adjacency matrix is not going to change within one timestep.
This allows for a trade-off between sufficient sparsity of $A$ while ensuring that all objects that potentially interact have corresponding entries in $A$.

In the experiments in Sec.~\ref{app:exp:adjacency}, we investigate the influence on the prediction performance for multiple different ways of predicting/using the adjacency matrix.

\subsection{Actions}\label{app:actions}
So far, the way we have formulated the graph neural network dynamics model in Sec.~\ref{app:propNets} does not contain a notion of actions.
Instead, we interpret an action as a modification to a node in the graph and train the GNN to predict the state of the nodes at the next time step as a result to this modification.
This allows us to not explicitly distinguish between controlled and uncontrolled/passive objects.

In order to realize modifications to a node and hence to incorporate actions in the first place, we exploit the fact that our object encoder is an implicit function of 3D world coordinates.
Assume that the object $j$ is articulated by a known rigid transformation $q\in\mathbb{R}^7$, which is the action.
As described in Sec.~\ref{app:novelSceneGeneration}, via \eqref{eq:zRigidTransormation} we can transform the object's latent vector $z_j^t$ into the transformed $\bar{z}^t_j =  z_j^{t+1} = \mathcal{T}(q)[z_j]$, which is kept constant during the propagation step of \eqref{eq:nodePropNetwork}, i.e.\ controlled nodes are excluded from the dynamics prediction, as their evolution is known through $\bar{z}^t_j$.

\subsection{Quasi-Static Dynamics}\label{app:quasiStatic}
If we assume quasi-static dynamics, meaning that the next system state only depends on the current latent state $z_{1:m}$ without history and immediate actions (refer to the discussion in the last paragraph Sec.~\ref{app:actions} about the notion of actions in this work), we can further increase the long-term stability by utilizing the adjacency matrices estimated by the learned model.
When an object is not involved in any interactions with other objects, then, under quasi-static assumptions, it does not change between time steps, i.e.\ its latent vector stays constant, which means we can set
\begin{align}
z_i^{t+1} = z_i^t~~~~\text{if}~~~~\forall_{j=1,\ldots,m,~ j\neq i} ~:~A_{ij} = 0.\label{eq:quaisStaticPropagation}
\end{align}
The condition $\forall_{j\neq i}~:~A_{ij} = 0$ means that the node associated with $z_i$ in the graph has no incoming edges.
As we will show in the experiments (Sec.~\ref{app:exp:adjacency}), this can greatly increase the stability for long-term open-loop model predictions as it will prevent drift in objects that do not take part in any interaction with other objects.

\subsection{Training}
We first train the compositional NeRF auto-encoder framework on training data, which gives us a dataset of trajectories of latent vectors.
The GNN dynamics model is then trained on the one-step mean squared error between $z^{t+1}_{1:m}$ and $z^{t}_{1:m}$ of samples of such trajectories using the ADAM optimizer.

Importantly, training the dynamics model does not require a dataset containing the actions.
It is sufficient to have video sequences and knowledge which object was the articulated one.
At inference time, one can also choose different objects to apply actions to in terms of rigid transformations.

\subsection{Forward Prediction Algorithm}\label{app:algoForwardPred}
Algorithm \ref{algo:forwardPred} summarizes the forward prediction procedure. The algorithm is visualized in Fig.~\ref{fig:forwardPredDiagram}.
\begin{algorithm}[t]
	\caption{Forward Prediction Model}
	\begin{algorithmic}[1]
		\State\textbf{Input:} Initial observation of the scene in terms of $I^{1:V}$, $K^{1:V}$, $M^{1:V}_{1:m}$, action sequence $q_{1:T}$, index $a$ of the articulated object
		\State $z_{1:m}^1 = \Omega\left(I^{1:V}, K^{1:V}, M_{1:m}^{1:V}\right)$\label{algo:forwardPred:Omega}
		\ForAll {$t=1, \ldots, T$}
		\State $\prescript{0}{}{z_{1:m}}^{t+1} = z_{1:m}^{t}$
		\State $\prescript{0}{}{z_{a}}^{t+1} = \bar{z}^t_a =  \mathcal{T}(q_t)[z_a^t]$\label{algo:forwardPred:q}
		\ForAll {$l=1, \ldots, L$}
		\State $\forall_{i,j, ~i\neq a}:~\prescript{l}{}{A}_{ij}^t = \prescript{l}{}{A}_{ij}^t\left(\prescript{l}{}{z_i}^{t+1}, \prescript{l}{}{z_j}^{t+1}\right)$ \label{algo:forwardPred:A}
		\State $\forall_{i,j~i\neq a}:~\prescript{l}{}{e}_{ij}^{t} = F_e\left(\prescript{l}{}{z_i}^{t+1}, \prescript{l}{}{z_j}^{t+1}\right)$
		\State $\forall_{i\neq a}\prescript{l}{}{z_i}^{t+1} = F_z\Big(z_i^t, \sum_{j ~:~ \prescript{l-1}{}{A}_{ij}^t = 1} \prescript{l-1}{}{e}_{ij}^{t}\Big)$
		\EndFor
		\State $z^{t+1}_i = \prescript{L}{}{z}_i^{t}$
		\State $z_i^{t+1} = z_i^t~~~~\text{if}~~~~\forall_{j=1,\ldots,m,~ j\neq i} ~:~A_{ij} = 0$\label{algo:forwardPred:exclude}
		\State Use $z^{t+1}_{1:m}$, e.g.\ render it from arbitrary views
		\EndFor
	\end{algorithmic}
	\label{algo:forwardPred}
\end{algorithm}
Starting from a single initial observation of the scene in terms of the images $I^{1:V}$ from $V$ many views and the objects masks $M^{1:V}_{1:m}$ of the $m$ many objects, Algorithm \ref{algo:forwardPred} predicts the latent vectors $z_{1:m}^t$ at times $t=1,\ldots,T$ for all objects in the scene given a desired action sequence $q_{1:T}$ of rigid transformations applied to object $a$.
At every point $t$ in time, the scene can be rendered from arbitrary view points from the predicted $z_{1:m}^{t}$.
Note that the masks of the objects are required only for the initial observation, i.e.\ no mask prediction has to be performed as compared to \cite{xu2020learning}.

In line \ref{algo:forwardPred:Omega}, the initial object encodings from the scene observation are computed.
Line \ref{algo:forwardPred:q} applies the action to the object with index $a$.
Lines 6-10 then perform the prediction step of the GNN in the latent space using message passing.
Crucially, in line \ref{algo:forwardPred:A}, the adjacency matrix is estimated from the current predictions during message passing (Sec.~\ref{app:adjacency}).
Note that here the original collision integral \eqref{eq:collisionIntegral}, computed on the grid $\mathcal{X}_h$, can be used without enlarging the intermediate occupancy grids, since, if during the message passing step objects interact that previously did not, it will be captured, as $A$ is estimated in every step of the message passing part. This leads to further increased prediction stability, as we will show in the experiments.
Finally, in line \ref{algo:forwardPred:exclude}, objects that have not interacted with other objects as predicted by the adjacency matrix are kept at their previous latent state (quasi-static assumption from Sec.~\ref{app:quasiStatic})

\section{Planning with Latent space RRT}\label{app:planningRRT}
In this section, we propose a planning algorithm to manipulate objects to achieve a desired goal using our scene encoding and dynamics model framework.
Note that planning and control is not the main focus of this work, however, the algorithm still contains important insights.

The main part of the planning algorithm is an RRT in the latent space.
Such latent space RRTs have been considered, for example, in \cite{ichter2019latentSpaceRRT}.
One central question here is how one can sample in the latent space effectively, since a uniform random sample in the latent space not necessarily is a valid (and/or uniform) sample in the original space.
In \cite{ichter2019latentSpaceRRT}, they assume to have access to a set of valid latent vectors from which they can sample.
In contrast, we can produce valid samples in the latent space directly by exploiting the properties of our model.

On a high level, our model iteratively perceives multi-view images of the scene, finds a plan using a Latent-Space RRT (LS-RRT) based on the forward predictions of the model over a long horizon, and then executes the found plan with Model-Predictive Control (MPC)~\cite{camacho2013model} for a shorter horizon.
We describe the algorithm here with pushing scenarios as considered in the experiments in mind.

\subsection{Planning}

Algorithm \ref{algo:rrt} summarizes the LS-RRT algorithm. We grow a tree in the latent space, starting at the latent vector $z^0 = z_{1:m}^0$ that represents the current state of the environment, encoded by the implicit object encoder $\Omega$ from the current visual observation of the scene.
In standard RRTs, a target is uniformly sampled in the configuration space to steer the growth of the tree towards a Voronoi bias.
To introduce a particular goal-targeted sampling bias and as we do not have an inverse model or steering function, we modify the standard approach as follows:

In a latent space RRT, sampling uniformly in the latent space does neither guarantee that the samples are from the latent space manifold nor that they explore the original space.
Therefore, we sample random targets $g\sim\mathcal{G}_\sampler$ not in the latent space directly, but only in the space of center of mass configurations of all objects, which is of dimension $2m$ in the experiments (objects and pusher).
In this way, we can design a sampling distribution $\mathcal{G}_\sampler$ biased to target configurations that have low costs, i.e., more objects within the goal region, or targets in which the articulated object (the pusher in the experiments) is close to one of the objects, inducing a bias for interaction.
This sampling distribution and cost evaluation $\mathcal{C}_g$ is possible because we can apply rigid transformations to the objects through our object encoder being an implicit function, since, for a sampled random target, we have to move the objects to this target to check the cost on the transformed configuration.
Further, the metric $d$ to select the expanded node is the $L_2$-norm in the full configurations between $g$ and the centers-of-masses computed from $z$. Using the predictions of the NeRF model, we can estimate (under homogeneous density assumption) the center of mass of an object with latent vector $z_j$ as
\begin{align}
x_j^\text{com}(z_j) = \frac{\int_{\mathcal{X}}x\cdot[\sigma(x, z_j) > \kappa]\diff x }{\int_{\mathcal{X}}[\sigma(x, z_j) > \kappa]\diff x},\label{eq:COM}
\end{align}
i.e.\ $d(z,g) = \left\|x_{1:m}^\text{com}(z) - g\right\|_2$.
Note that the sampling distribution, cost function evaluation and metric calculation are done solely based on predictions of the model.
At no point the model has access to ground truth center-of-mass information.

Finally, as we do not have an inverse dynamics model or an other kind of steering function, we expand the tree using a random action $q$, similar to control trees. However, our goal-targeted node selection ensures that the tree expands effectively.

\begin{algorithm}[t]
	\caption{Latent-Space RRT}
	\begin{algorithmic}[1]
		\State\textbf{Input:} Initial latent vector $z^0$, action space $\{q\}$, metric $d$, ()sparse) goal cost function $\mathcal{C}_g$.
		\State Initialize tree $\mathbb{T}$ with $z^0$ as the root.
		\While {time remains}
		\State Sample a target $g \sim \mathcal{G}_\sampler$ 
		\State Find nearest $z_{\text{prop}}=\arg\min_{z\in \mathbb{T}}d(z, g)$
		\State Sample action $q$ randomly
		\State Execute action $q$ in learned forward model with Algorithm \ref{algo:forwardPred} to get new latent vector $z_\text{next}$ 
		\State Add $z_{\text{next}}$ into the tree $\mathbb{T}$, record action $q$ and its parent $z_\text{prop}$
		\If {goal $\mathcal{C}_g(z_\text{next})$ fulfilled}
		\State \Return the action sequence from $z^0$ to $z_\text{next}$
		\EndIf
		\EndWhile
	\end{algorithmic}
	\label{algo:rrt}
\end{algorithm}

\subsection{Cost-Functions}
As our decoder is based on NeRFs, our method is able to provide a lot of flexibility in defining cost functions.
At any time instance forward predicted by the dynamics model, we can render an image from an arbitrary view or reconstruct the objects in 3D.
This enables, for example, to have the following options to define cost functions for planning:
\begin{itemize}
	\item Loss on image rendered from arbitrary views (with known camera matrix).
	\item Loss on density prediction of the NeRF reconstruction.
	\item Loss on point-cloud reconstructed from $\sigma_j$ (including color information on each point).
	\item Loss on center-of-mass predicted by the model via \eqref{eq:COM}.
\end{itemize}
All these options can be defined for specific objects, the whole scene, or anything in between.

\subsection{Model-based Control}

Although our model achieves impressive performance over a long horizon, the accumulated prediction errors may still lead to a failure when executing the plans open-loop.
We therefore apply an MPC scheme, which in each cycle feeds the current visual observation into the model, samples and select actions that match the plan (in terms of the center-of-mass metric) within a short horizon predicted by the learned dynamics model.
If there is a significant mismatch between the plan and current observation, the LS-RRT is used again to find a new long-term plan starting from the current observation.

\clearpage

\section{Experiments -- Simulation}
\begin{figure*}
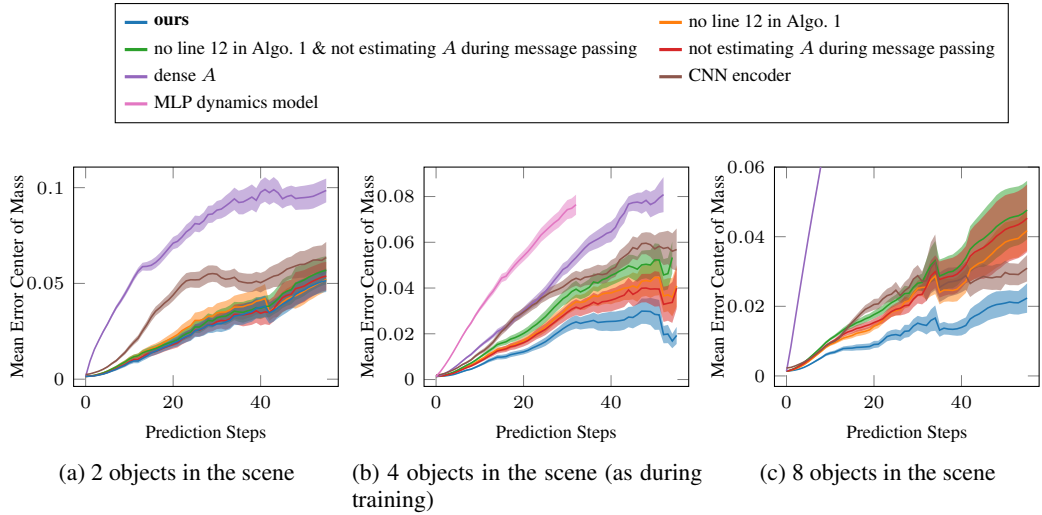

	\centering
	\scalebox{1.0}{\hspace{1.1cm}\input{plots/comLegend}}\\
	\subfloat[2 objects in the scene]{%
		\input{plots/com_nM_2}%
		\label{fig:com:2}
	}%
	\subfloat[4 objects in the scene (as during training)]{%
		\input{plots/com_nM_4}%
		\label{fig:com:4}
	}%
	\subfloat[8 objects in the scene]{%
		\input{plots/com_nM_8}%
		\label{fig:com:8}
	}%
	\caption{Performance comparison for simulation experiments in terms of mean center of mass prediction error of the objects over the number of prediction steps into the future on test data containing different numbers of objects. 
		The center of mass is estimated from the density prediction of the NeRF corresponding to each object according to \eqref{eq:COM}.
		One prediction step corresponds to a 2 cm movement of the pusher, i.e. for 50 steps the pusher has moved 1 m. For the MLP dynamics model in (b), it at some point did not predict all objects anymore, which meant no center of mass could be calculated further. Our method outperforms all baselines.
	}
	\label{fig:com}
\end{figure*}

In the simulated experiments, we focus on a pushing task in scenarios with multiple box-shaped objects on a table, see, e.g.\, Fig.~\ref{fig:sceneBB} or \ref{fig:trainingMasks} for such scenes.
For a quantitative analysis and comparison to multiple baselines, we investigate the forward prediction error of the model both in the image space (Fig.~\ref{fig:baselinesImage}) and, for baselines that use a compositional NeRF decoder, the error in predicting the center of mass of the objects (Fig.~\ref{fig:com}) over long-horizons.
In all plots of Fig.~\ref{fig:baselinesImage} and Fig.~\ref{fig:com}, the blue curve corresponds to our proposed framework as summarized in Algo.~\ref{algo:forwardPred}.
Sec.~\ref{app:exp:boxSorting} presents planning and execution results for a challenging box sorting task.

Please refer to the supplementary material for videos showing the reconstructions of the model, forward predictions, novel scene generation, and planning/execution results.

\subsection{Setup}\label{sec:exp:setup}
\begin{figure}
	\centering
	\includegraphics[trim={3cm 4cm 3cm 2cm}, clip, width=4.5cm]{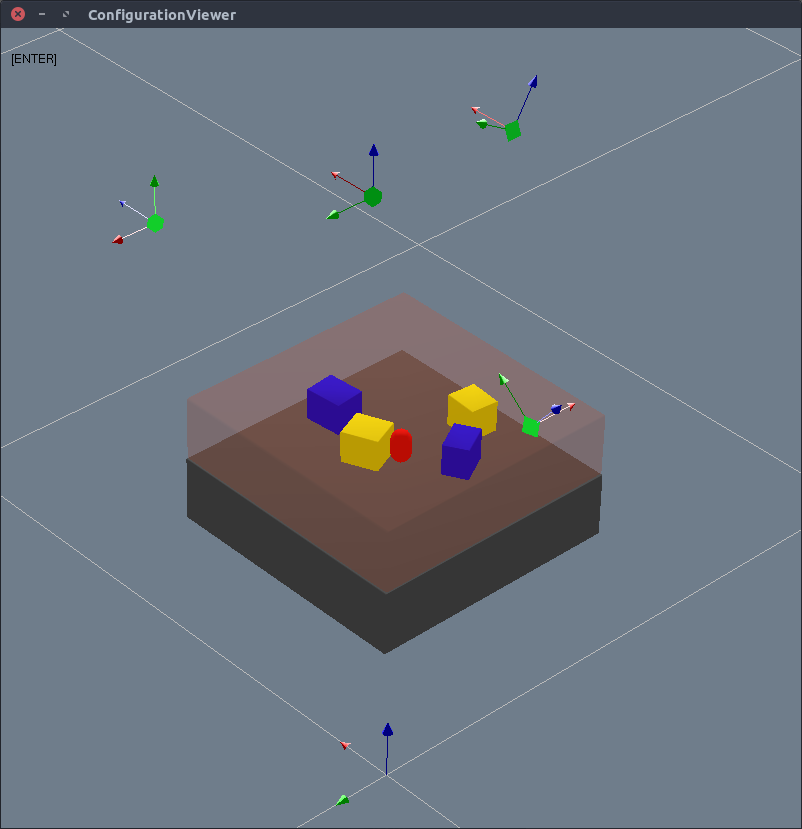}
	\caption{Example scene. The red shaded box visualizes the workspace set volume $\mathcal{X}$. The green coordinate systems denote the camera origins and view directions. The red cylinder is the pusher that is actuated.}
	\label{fig:sceneBB}
\end{figure}
We consider a rigid-body scenario with multiple objects on a table, see Fig.~\ref{fig:sceneBB} for an example.
In all cases, the red cylinder is the pusher that is articulated in order to push the other objects around.

This scenario is challenging due to multiple reasons.
First, it is composed of many objects, which implies not only a broad scene distribution, but especially also that many objects can interact.
The mechanics of such multi-body pushing is non-trivial, since, for instance, contact can be established and broken between the objects at multiple phases of the motion. Contact between multiple objects at the same time can occur.
Furthermore, we do not assume that the red pusher starts in contact with an object. 
Hence, if a task implies that an object should be pushed, long-term predictions inherently have to be made in order to establish contact, before any object movement is registered.

The workspace is an area of 40 cm $\times$ 40 cm $\times$ 10 cm and we choose $\mathcal{X}_h\in\mathbb{R}^{10\times40\times40}$, i.e.\ a resolution of 1 cm. 

All scenes in the training data contain 4 box-shaped objects of randomly sampled sizes, positions and orientations (5 dimensional parameter space for each of the 4 objects) and one cylinder-shaped object with randomly sampled position.

To generate the training data, we randomly sample one of the 4 objects and then move the red pusher towards the center of this chosen object (with Gaussian noise added to the direction vector in each time step) until either the pusher leaves the workspace, in which case a new target object is chosen, or an object is pushed outside the workspace, in which case the data collection for this scene is terminated and a new scene is sampled.
In total, the training dataset contains 5752 scenes with an average sequence length of 17.
We generate 3 test datasets for evaluating the reconstruction and prediction performance which contain 2, 4, and 8 objects, respectively, plus the pusher.
There are 312 scenes for each test dataset, generated with a different random seed than the training data.
As visualized in Fig.~\ref{fig:sceneBB} by the green coordinate systems, we choose 4 camera views for each scene.

\begin{figure*}
	\centering
	\subfloat[Reconstructions on observed views]{
		\centering
		\begin{tabular}{@{}c@{\hspace{0.1cm}}c@{\hspace{0.1cm}}c}
			\multicolumn{3}{c}{\scriptsize ground truth, observed views} \\
			\includegraphics[width=2.1cm]{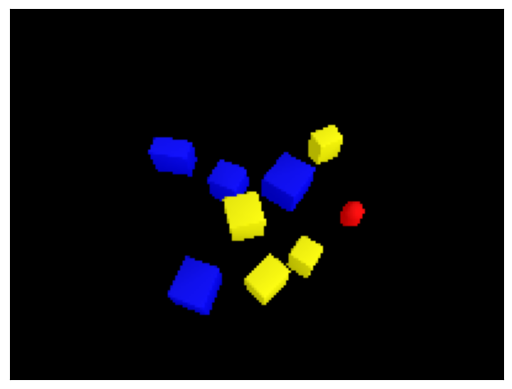}&
			\includegraphics[width=2.1cm]{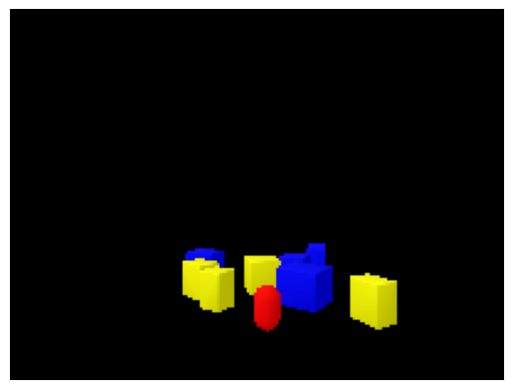}&
			\includegraphics[width=2.1cm]{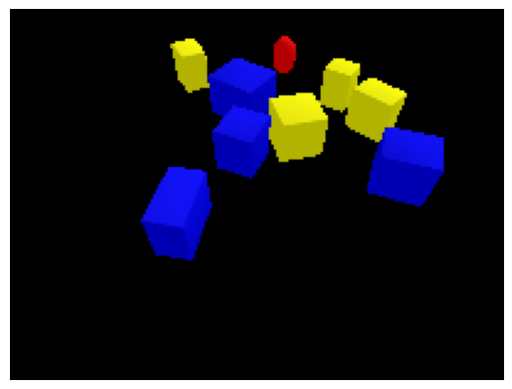}\\
			\multicolumn{3}{c}{\scriptsize reconstruction by our model} \\
			\includegraphics[width=2.1cm]{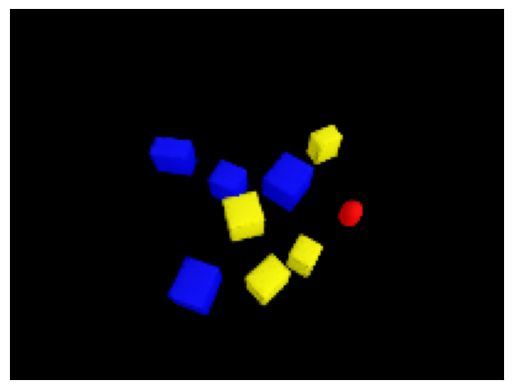}&
			\includegraphics[width=2.1cm]{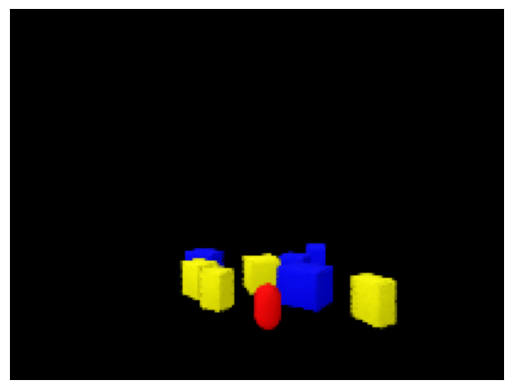}&
			\includegraphics[width=2.1cm]{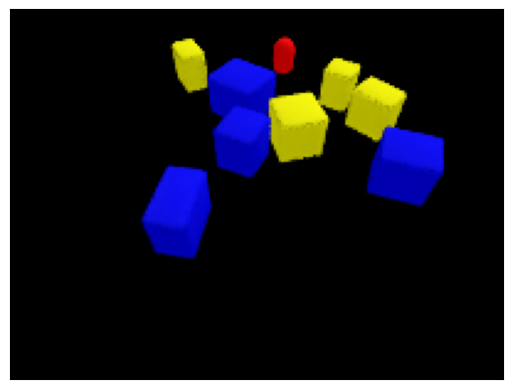}
		\end{tabular}
		\label{fig:eightObjectsApp:observed}
	}
	\subfloat[Novel view synthesis from latent vectors computed from the views in (a)]{
		\centering
		\begin{tabular}{c@{\hspace{0.1cm}}c@{\hspace{0.1cm}}c}
			\multicolumn{3}{c}{\scriptsize ground truth, novel views} \\
			\includegraphics[width=2.1cm]{images/eightObjects/withMask5.png}&
			\includegraphics[width=2.1cm]{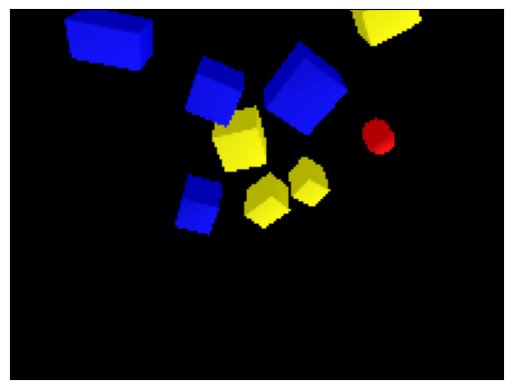}&
			\includegraphics[width=2.1cm]{images/eightObjects/withMask6.png}\\
			\multicolumn{3}{c}{\scriptsize reconstruction for novel views based on views of (a)} \\
			\includegraphics[width=2.1cm]{images/eightObjects/reconstruction5.png}&
			\includegraphics[width=2.1cm]{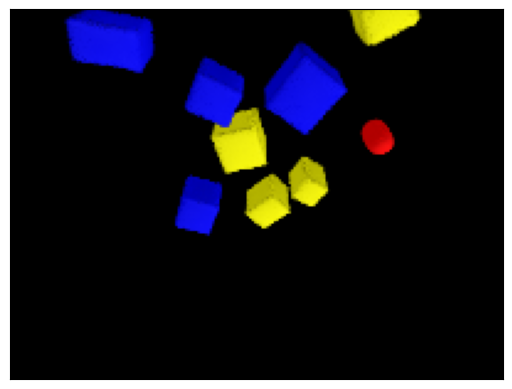}&
			\includegraphics[width=2.1cm]{images/eightObjects/reconstruction6.png}
		\end{tabular}
		\label{fig:eightObjectsApp:novelView}
	}
	\caption{Generalization to more objects (8 objects plus 1 pusher) than during training. The scenes in the training dataset contain exactly 4 objects and one pusher. The top row in (a) corresponds to the observed scene from different views; the bottom row is the reconstruction by our method. The bottom row in (b) is the reconstruction of novel views (top row ground truth) with the latent vectors computed from the views in the top row of (a).
	}
	\label{fig:eightObjectsApp}
\end{figure*}

\subsection{Importance of Estimating the Adjacency Matrix}\label{app:exp:adjacency}
This section provides a more detailed investigation of the importance of the adjacency matrix than in the main text, where we only have discussed a dense adjacency matrix. 

In Sec.~\ref{app:adjacency}, we have proposed how the adjacency matrix of the GNN can be estimated from the density predictions of the learned NeRFs and that under quasi-static assumptions this estimated adjacency matrix can further be exploited to increase the long-term stability of the predictions, cf.~Sec.~\ref{app:quasiStatic}.
Here we investigate the consequences of utilizing the adjacency matrix this way by comparing the full Algorithm \ref{algo:forwardPred} to the following three ablations.
In all these ablations, the rest of the method remains the same, i.e.\ same object encoder, same GNN, same compositional NeRF decoder.

\paragraph{Not exploiting quasi-static assumption}
In this case, line \ref{algo:forwardPred:exclude} of Algorithm \ref{algo:forwardPred} is not used, i.e.\ the latent vectors of all objects, even when they do not interact with other objects as estimated through the model, are updated using the model forward predictions.

\paragraph{Adjacency matrix estimation not during message passing}
Here, we estimate the adjacency matrix only at the beginning of the message passing step, i.e.\ before line 6 in Algorithm \ref{algo:forwardPred}.
In order to ensure that it can still capture all object interactions that might occur during the message passing step, we enlarge the determined occupancy grids exactly the same way as for training, see the discussion in Sec.\ref{app:adjacency}.
The effects of this are that objects that are close to each other but do not interact still have entries in $A$ indicating that they interact, which means slight errors in the predictions accumulate and lead to drift, although the object would not move in reality.

\paragraph{Dense adjacency matrix}
We further consider a dense adjacency matrix, i.e.\ where the network has to figure out from the latent vectors themselves if objects interact.
Preventing drift in this case is considerably harder.

\paragraph{Results adjacency matrix}
In Fig.~\ref{fig:com} one can see the mean error of the model predicting the center of mass, computed from its density predictions of the NeRFs for each object according to \eqref{eq:COM}, over the number of steps predicted into the future on the test dataset for different numbers of objects in the scene.

As one can see in Fig.~\ref{fig:com:2}, for the two object case, the choices of how the adjacency matrix is used, as long as it is not a dense one, are not significant.
For the 4 (Fig.~\ref{fig:com:4}) and 8 (Fig.~\ref{fig:com:8}) object case, however, our proposed utilization of the adjacency matrix, i.e.\ estimating it during propagation steps and using it to exploit the quasi-static assumption, leads to a significant increase in performance.
This can be explained by the fact that utilizing the adjacency matrix as we propose leads to significantly less drift.
Especially with the dense adjacency matrix, the predictions are very unstable for all, the 2, 4, and 8 object case.
Fig.~\ref{fig:forwardPred3Objects:denseA} shows this qualitatively.
In the 8 object case, the predictions with the dense $A$ are basically useless after only a few time-steps, showing that it has overfitted to the number of 4 objects as in the training data.

\subsection{Advantages of Implicit Object Encoder -- Comparison to CNN Encoder}
This experiment has already be mentioned in the main text, but here we provide more details.
We exchange the implicit object encoder with a 2D CNN object encoder.
The resulting auto-encoder framework is very similar to the architecture of \cite{stelzner2021decomposing}.
More specifically, we encode each masked image observation with a 2D CNN to produce a feature vector. The feature vectors from the different views are aggregated into the final latent vectors for each object.
We use the encoder architecture from \cite{li20223d}, but adjust it to the compositional multi-object case by incorporating object masks.
Since this encoder is not an implicit function of $\mathcal{X}$, we cannot modify the latent vectors by applying rigid transformations and hence need to encode the actions differently.
In order to do so, we train a separate MLP network that predicts the latent vector of the pusher resulting from applying an action to it.
This gives the modified $z_a$ for line \ref{algo:forwardPred:q} in algorithm \ref{algo:forwardPred} for the CNN encoder baseline.
The rest of the architecture, i.e.\ the GNN, the compositional NeRF decoder, estimating the adjacency matrix during message passing from the model, etc., stays the same.

As can be seen in Fig.~\ref{fig:com:2} and Fig.~\ref{fig:baselinesImage}, replacing the proposed implicit object encoder with a CNN encoder, the performance is better compared to the other baselines, but still clearly worse than with the proposed method.

\subsection{Planning and Execution Results on Box Sorting Task}\label{app:exp:boxSorting}
To demonstrate the effectiveness of the learned model, we utilize it to solve a challenging box sorting task, where the red pusher needs to push the blue and yellow boxes into their corresponding goal regions as shown in Fig.~\ref{fig:boxPlanning}.
This task is in part inspired by the object sorting task in \cite{pmlr-v164-florence22a}.
The cost function $\mathcal{C}_g$ in Algorithm \ref{algo:rrt} determines how many objects are outside of their goal region, which is computed for each object $j$ from their corresponding density $\sigma_j$ and color $c_j$ predictions of the model itself. The goal is fulfilled if all objects are in their respective goal regions. 

This object-sorting task is challenging for multiple reasons.
First, the dynamics of pushing is non-trivial \cite{hogan2016feedback, zhou2019pushing, 21-driess-IJRR, schubert2021learning}.
In our particular case, many objects potentially interact, which further complicates the setup.
Pushing one object could undo an object that is already at the goal, hence a greedy strategy of just pushing the objects straight to the goal region would fail.
In addition, movements, i.e.\ actions, of the pusher do usually not immediately lead to a change in the cost function, since contact with the object from a suitable side has to be established, for which it is often necessary for the pusher to move around objects \cite{driess2022CoRL, schubert2021learning}.
Therefore, applying planning methods that are too local like a cross-entropy method would fail for this scenario.
Our prediction model combined with the LS-RRT algorithm solves these tasks efficiently, just from image observations of the scene, see Fig.~\ref{fig:firstPage} and the video.   
In the first row of Tab.~\ref{tab:plan_res}, we show the total size of the exploration trees for solving the tasks for scenes that contain $1$ to $6$ objects.

As a baseline comparison, we consider planning with a GNN that uses a fully connected adjacency matrix (Sec.~\ref{app:exp:adjacency}) to understand the importance of a precise dynamics model.
From the second row of Tab.~\ref{tab:plan_res}, one can see that planning with a dense $A$ either fails to find a solution as the pusher may not be able to move the object correctly, or generates a plan that is tough to follow in the simulation environment.
The reason for this is that with a dense $A$, as shown in Fig.~\ref{fig:forwardPred3Objects:denseA}, the model induces too much drift of objects that do not interact, which makes planning for pushing scenarios extremely difficult.
Objects can also drift closer or away again to/from the goal by model errors.

Finally, we compare the LS-RRT with a naive control tree algorithm, where we still use our full model, but only sample random nodes in the tree to extend.
As shown in the last row of Tab.~\ref{tab:plan_res}, though the planner can find a path to push a single object, it fails to solve tasks containing more objects within the timeout.
This demonstrates the benefits of our object encoder being an implicit function and being able to relate information in the 3D world in terms of center of mass predictions via the learned NeRFs to the latent vectors, both of which make planning much more efficient compared to naive control trees.

\begin{table}
	\caption{Number of samples to find a solution with the latent space RRT planning algorithm for box-sorting scenarios containing 1, 2, 4, and 6 objects. NS means no solution found within $10^5$ samples. FE means failed to execute the found plan for $10$ times.}
	\centering
	\small
	\begin{tabular}{lllll}
		\toprule
		Number of Objects   &1  & 2  & 4  & 6 \\
		\midrule
		RRT with full model & 256 & 2341 & 23819 & 85022\\
		RRT with dense $A$ & NS & FE & FE & NS\\
		Control tree with full model & 24019 & NS & NS & NS\\
		\bottomrule
	\end{tabular}
	\label{tab:plan_res}
\end{table}

\clearpage

\section{Experiments -- Real World}
\vspace{-0.1cm}
In the real world experiments, we consider a pushing scenario with different objects on a table.
The pusher (blue) is articulated by a Franka Emika Panda robot.
All real world experiments use the same 4 camera views (see Fig.~\ref{fig:realWorldSetup}).
As one can see, these 4 cameras are side-views, i.e.\, in particular, no top-down view is available.
Obtaining top-down views in such a setup is challenging, as the robot obstructs the objects in a top-down view.
We show in the video renderings from the learned NeRF model from top-down views.
We use Intel Realsense cameras.
To obtain the object masks in each view, we employ a simple color thresholding method.

The training data consists of image sequences from the different views, only.
In particular, the movements of the robot, i.e.\ the rigid transformations applied to the blue pusher by the robot, are not required to be available.

\subsection{(Rigid) Objects}\label{app:exp:real:rigid}
\begin{figure}[!b]
	\centering
	\includegraphics[width=9cm]{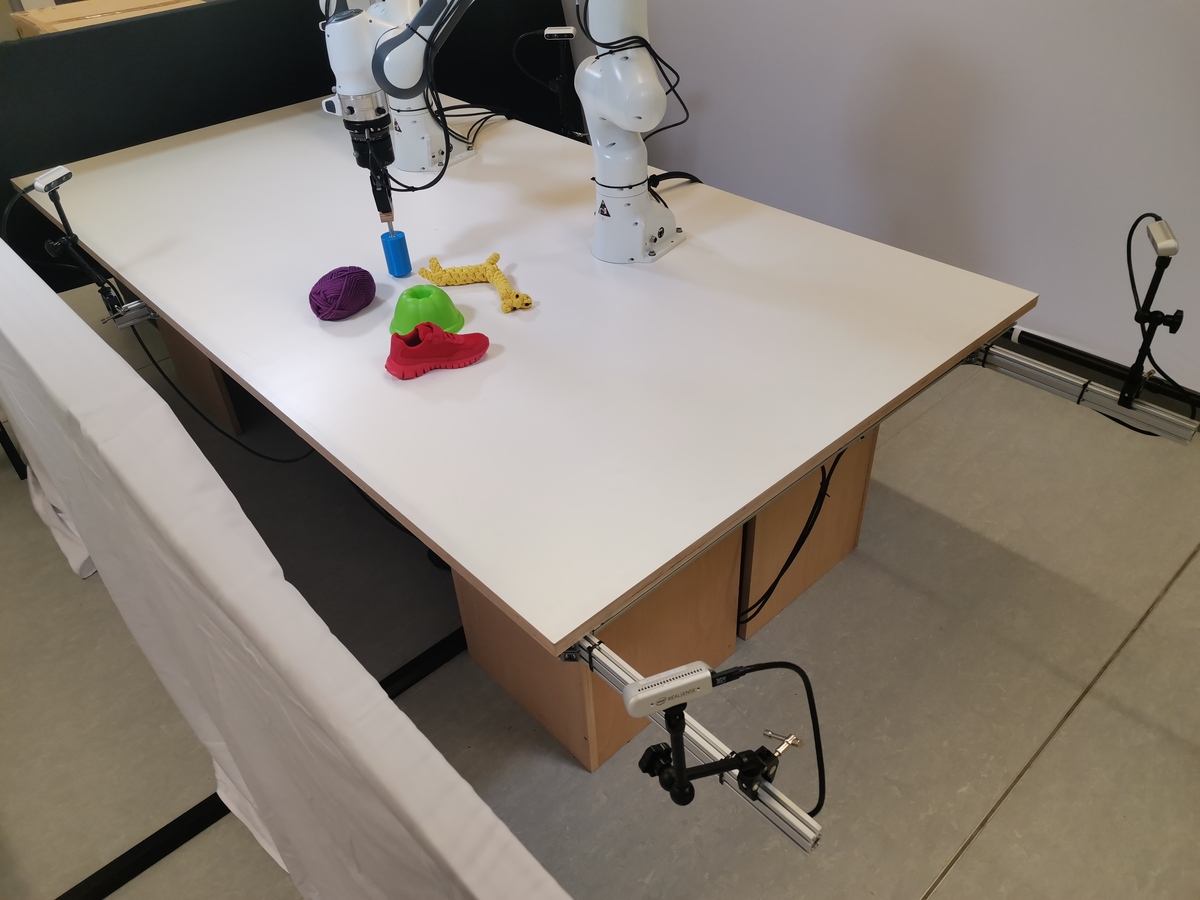}
	\caption{Setup of real world experiments. 4 camera views from the side. No top down view.}
	\vspace{-0.3cm}
	\label{fig:realWorldSetup}
\end{figure}

\vspace{-0.1cm}
In this experiment, we consider 4 objects, a shoe (red), a giraffe-shaped toy (yellow), a sand mold (green), and a ball of wool (violet).
Although these objects are not rigid, they mainly behave as rigid objects when being pushed.
See Fig.~\ref{fig:realWorldSetup} for these objects.

We train the framework on a dataset of 2000 random pushes in total, which takes about 8 hours to collect.
We randomly apply push directions with a simple heuristics to try to prevent that the objects are being pushed outside of the workspace.
Further, this heuristic biases the random push direction sampling to a randomly chosen object, for which we use the depth information of one of the Realsense cameras.
Note that except for this heuristic to collect the data more efficiently (otherwise pure random pushes would interact with the objects much more rarely), the depth information from the Realsense cameras is \emph{not} used.

Half of the training scenes contain the giraffe and the shoe, the other half the shoe, the sand mold, and the ball of wool.
The giraffe is never in one scene together with the ball of wool.
In the video, we show that the model is still able to reasonably predict the dynamics of a scene containing a giraffe and a wool, even when they interact.

\begin{figure}
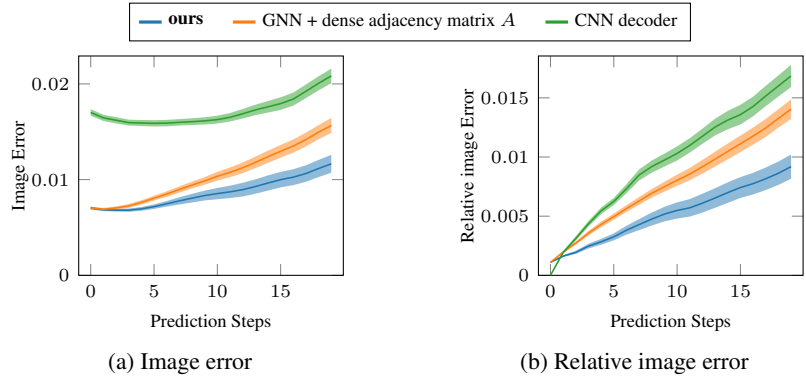

	\centering
	\vspace{-0.5cm}
	\input{plots/real/legend}\\[-0.2cm]
	\subfloat[Image error]{
		\input{plots/real/woolSandtoyShoe_predTrue}
	}
	\hspace{1cm}
	\subfloat[Relative image error]{
		\input{plots/real/woolSandtoyShoe_predRecon}
	}
	\caption{Performance on real world test scenes containing a shoe, a sand mold, and a ball of wool. The training data also contained scenes with these objects. Quantitative performance comparison in terms of image error over number of prediction steps into the future. (a) image error between prediction and ground truth. (b) image error of forward prediction relative to image reconstructed when observations at each time step are available.}
	\label{fig:realWorldWoolSandtoyShoe}
\end{figure}

\begin{figure}
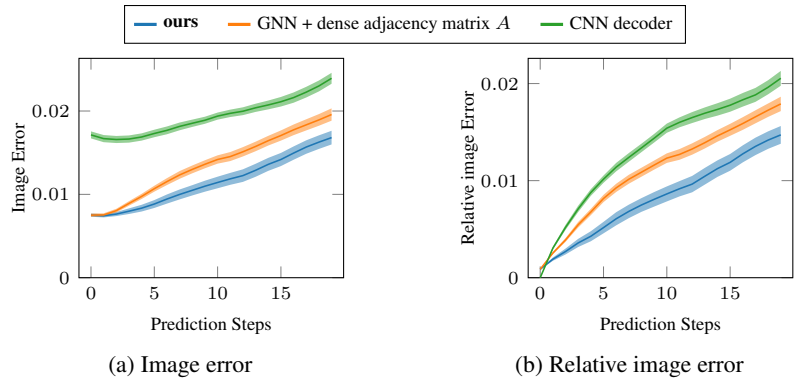

	\centering
	\input{plots/real/legend}\\[-0.2cm]
	\subfloat[Image error]{
		\input{plots/real/shoeGiraffe_predTrue}
	}
	\hspace{1cm}
	\subfloat[Relative image error]{
		\input{plots/real/shoeGiraffe_predRecon}
	}
	\caption{Performance on real world test scenes containing a giraffe and a shoe. The training data also contained scenes with these objects. Quantitative performance comparison in terms of image error over number of prediction steps into the future. (a) image error between prediction and ground truth. (b) image error of forward prediction relative to image reconstructed when observations at each time step are available.}
	\label{fig:realWorldShoeGiraffe}
\end{figure}

Fig.~\ref{fig:realWorldWoolSandtoyShoe} and Fig.~\ref{fig:realWorldShoeGiraffe} show the performance in terms of the forward predicted image reconstruction error for test scenes from the same distribution as during training.
As one can see, our method outperforms the dense adjacency and CNN decoder baselines.

\begin{figure}
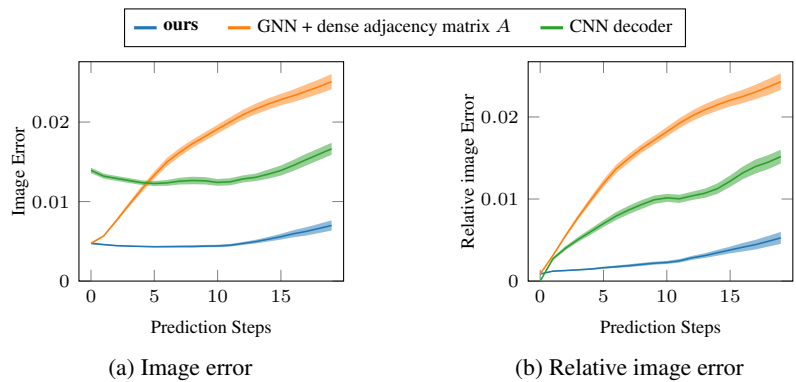

	\centering
	\input{plots/real/legend}\\[-0.2cm]
	\subfloat[Image error]{
		\input{plots/real/woolSandtoyShoe_woolSandtoy_predTrue}
	}
	\hspace{1cm}
	\subfloat[Relative image error]{
		\input{plots/real/woolSandtoyShoe_woolSandtoy_predRecon}
	}
	\caption{Performance on real world test scenes containing a sand mold and a ball of wool, i.e.\ this experiment shows generalization over less numbers of objects in the scene than during training. As one can see, our method has no difficulties in generalizing to a different number of objects in the scene than during training, while the baselines have more difficulties here than on test data containing the same number of objects as during training. Quantitative performance comparison in terms of image error over number of prediction steps into the future. (a) image error between prediction and ground truth. (b) image error of forward prediction relative to image reconstructed when observations at each time step are available.}
	\label{fig:realWorldWoolSandtoy}
\end{figure}

\begin{figure}
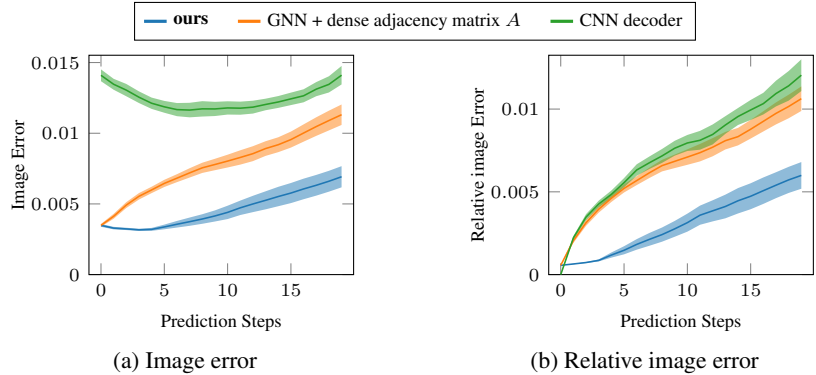

	\centering
	\input{plots/real/legend}\\[-0.2cm]
	\subfloat[Image error]{
		\input{plots/real/shoeGiraffe_shoe_1_predTrue}
	}
	\hspace{1cm}
	\subfloat[Relative image error]{
		\input{plots/real/shoeGiraffe_shoe_1_predRecon}
	}
	\caption{Performance on real world test scenes containing a shoe only, i.e.\ this experiment shows generalization over less numbers of objects in the scene than during training. As one can see, our method has no difficulties in generalizing to a different number of objects in the scene than during training, while the baselines have more difficulties here than on test data containing the same number of objects as during training. Quantitative performance comparison in terms of image error over number of prediction steps into the future. (a) image error between prediction and ground truth. (b) image error of forward prediction relative to image reconstructed when observations at each time step are available.}
	\label{fig:realWorldShoe}
\end{figure}

Fig.~\ref{fig:realWorldWoolSandtoy} and Fig.~\ref{fig:realWorldShoe} show the performance for scenes that contain a different number (in this case less) of objects than during training.
Our method achieves a very low error here, even lower as in Fig.~\ref{fig:realWorldWoolSandtoyShoe} and Fig.~\ref{fig:realWorldShoeGiraffe} where the same number of objects are in the test scenes.
In contrast, the dense adjacency matrix baseline has overfit to the number of objects in the scene and performs poorly, although the scenes might seem easier, as they contain less objects.

\vspace{-0.3cm}

%\clearpage
\subsection{Deformable Object}\label{app:exp:real:deformable}
\begin{figure}
	\centering
	\includegraphics[width=6cm]{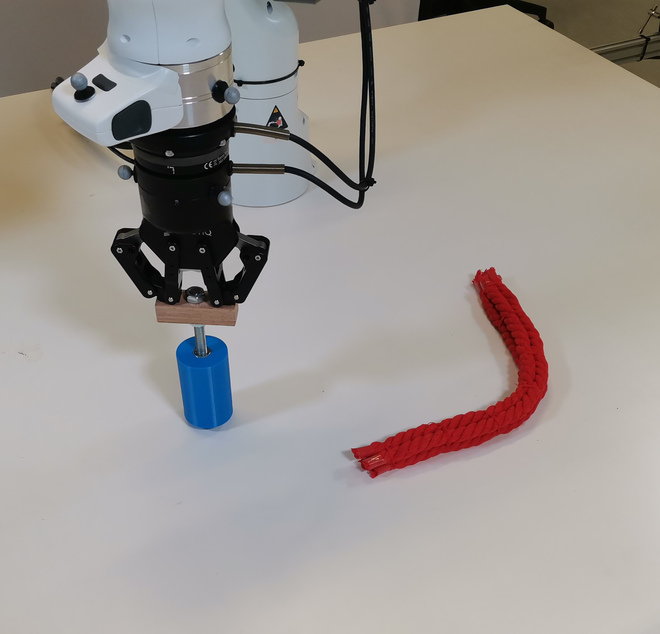}
	\caption{Real world experiment with deformable object -- a piece of rope (red).}
	\label{fig:realWorldDeformableSetup}
\end{figure}

Finally, we consider a real world experiment where a rope is pushed on a table.
This rope, visualized in Fig.~\ref{fig:realWorldDeformableSetup}, is deformable in the sense that after the interaction with the pusher it remains in its last, deformed state.

We train the framework on a dataset of 500 random pushes with the same method as in Sec.~\ref{app:exp:real:rigid}, which takes about 2 hours to collect.
No changes were required for applying our method to such a deformable object.

Please refer to the video for a visualization of the forward predictions of this experiment.

\begin{figure}
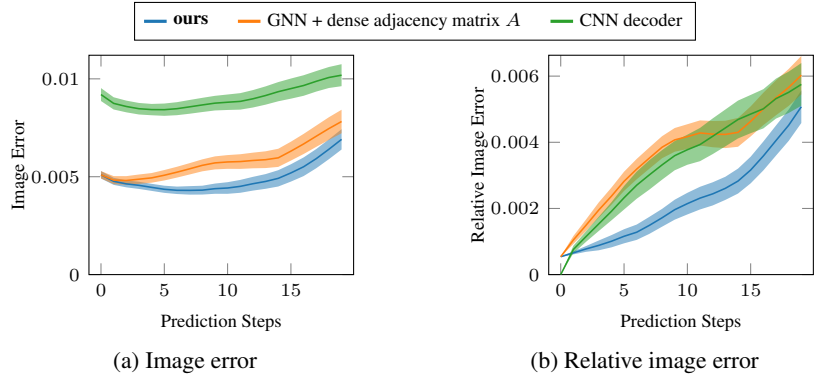

	\centering
	\input{plots/real/legend}\\[-0.2cm]
	\subfloat[Image error]{
		\input{plots/real/rope_predTrue}
	}
	\hspace{1cm}
	\subfloat[Relative image error]{
		\input{plots/real/rope_predRecon}
	}
	\caption{Real world deformable object experiment. Quantitative performance comparison in terms of image error over number of prediction steps into the future. (a) image error between prediction and ground truth. (b) image error of forward prediction relative to image reconstructed when observations at each time step are available.}
	\label{fig:realWorldDeformableComparison}
\end{figure}

Fig.~\ref{fig:realWorldDeformableComparison} compares on a test dataset containing 50 scenes the performance of our method with a version where the adjacency matrix is dense and where the decoder is a convolutional neural network instead of a compositional NeRF.
This shows that compositionality in our framework has a benefit even if the scene only contains one object and one pusher, as representing objects with individual latent vectors that parameterize individual NeRFs allows us to compute the adjacency matrix from the model's predictions adaptively, leading to better performance.

\section{Network Architectures}
The dimension of the latent vectors $z_j$ is $k=64$ for each object. All hidden activation functions are ReLUs.

The MLP that encodes the projected coordinate (see Fig.\ref{fig:omega:E}) of the implicit object feature encoder $E$ has one layer with output dimension 32.
The other MLP in $E$ has 2 hidden layers with 128 units each and an output dimension of $n_o = 64$.

The volumetric feature encoder $\Phi$ consists of three 3D convolutional layers with kernel size 3 and channel size 128, each. 
Layers 2 and 3 have strides of 2.
After the convolutional layers, the output is flattened and processed with 3 dense layers with 300 hidden units each.

The NeRF network $f$ first lifts the 3D input to 64 dimensions with an MLP, where it is concatinated with the latent vector $z$.
This is followed by 3 hidden layers with 300 units each.
For the density output $\sigma$, we use a softplus activation and a sigmoid for the color outputs $c$.

Both the edge encoder $F_e$ and the node propagator network $F_z$ have 3 hidden layers with 256 units each.

\end{document}